\documentclass[lettersize,journal]{IEEEtran}

\usepackage[T1]{fontenc}
\usepackage[utf8]{inputenc}
\usepackage{microtype}
\usepackage{textcomp}

\usepackage{amsmath,amssymb,amsfonts}
\usepackage{bm}
\usepackage{mathtools}
\usepackage{nicefrac}

\usepackage{graphicx}
\usepackage[caption=false,font=footnotesize]{subfig}

\usepackage[table]{xcolor}
\usepackage{array}
\usepackage{tabularx}
\usepackage{booktabs}
\usepackage{multirow}
\usepackage{makecell}

\definecolor{lightblue}{RGB}{173,216,230}
\definecolor{palepink}{RGB}{255,224,230}
\definecolor{orange}{RGB}{255,165,0}

\usepackage{algorithm}
\usepackage{algpseudocode}

\usepackage[nocompress]{cite}
\usepackage{hyperref}
\usepackage[capitalise,nameinlink]{cleveref}

\hypersetup{
    colorlinks = true,
    citecolor  = blue,
    linkcolor  = black,
    urlcolor   = black,
    filecolor  = black
}

\usepackage{comment}
\usepackage{relsize}

\usepackage{orcidlink}

\usepackage{stfloats}
\usepackage{balance}

\def\BibTeX{{\rm B\kern-.05em{\sc i\kern-.025em b}%
\kern-.08em T\kern-.1667em\lower.7ex\hbox{E}\kern-.125emX}}

\begin{document}

\title{A$^2$Safe: Counterfactual Evidence-Aligned Adaptive Agent Collaboration for Safe and Effective Visual Question Answering}

\author{
Quanxing~Xu\orcidlink{0009-0008-4354-8371},
Ling~Zhou\orcidlink{0000-0002-8313-5749},
Xian~Zhong\orcidlink{0000-0002-5242-0467},~\IEEEmembership{Senior~ Member,~IEEE},
Jinyu~Tian\orcidlink{0000-0002-2449-5277},~\IEEEmembership{Member,~IEEE},
Xiaohua~Huang\orcidlink{0000-0001-8897-3517},~\IEEEmembership{Senior~ Member,~IEEE},
Rubing~Huang\orcidlink{0000-0002-1769-6126},~\IEEEmembership{Senior~ Member,~IEEE},
and~Chia-Wen~Lin\orcidlink{0000-0002-9097-2318},~\IEEEmembership{Fellow,~IEEE}

\thanks{Manuscript received September 20, 2026. This work was supported in part by the Science and Technology Development Fund of Macau, Macao SAR, under Grants 0069/2025/RIB2 and 0021/2023/RIA1; the National Natural Science Foundation of China under Grant 62271361; the Hubei Provincial Key Research and Development Program under Grant 2024BAB039; and the Hubei Provincial Natural Science Foundation under Grant 2026AFB663. (\textit{Corresponding authors: Ling Zhou and Xian Zhong.})}

\thanks{Quanxing Xu, Ling Zhou, Jinyu Tian, and Rubing Huang are with the School of Computer Science and Engineering, Macau University of Science and Technology, Macao SAR 999078, China (e-mail: 3230002299@student.must.edu.mo; lzhou@must.edu.mo; jytian@must.edu.mo; rbhuang@must.edu.mo).}

\thanks{Xian Zhong is with the Hubei Key Laboratory of Transportation Internet of Things, School of Artificial Intelligence, Wuhan University of Technology, Wuhan 430070, China
(e-mail: zhongx@whut.edu.cn).}

\thanks{Xiaohua Huang is with the Oulu School, Nanjing Institute of Technology, Nanjing 210096, China (e-mail: xiaohuahwang@gmail.com).}


\thanks{Chia-Wen Lin is with the Department of Electrical Engineering, National Tsing Hua University, Hsinchu 30013, Taiwan (e-mail: cwlin@ee.nthu.edu.tw).}
}

\markboth{IEEE TRANSACTIONS ON MULTIMEDIA, 2026}%
{Xu \MakeLowercase{\textit{et al.}}: A$^2$Safe: Counterfactual Adaptive Evidence Alignment for Safe and Effective Visual Question Answering}

\maketitle

\begin{abstract}
Visual Question Answering (VQA) with Multimodal Large Language Models (MLLMs) requires not only producing safe and effective responses, but also grounding safety decisions in the multimodal evidence that determines risk. Recent safety-alignment methods improve refusal behavior and contextual risk awareness, yet correct safety outcomes may still rely on superficial textual or visual correlations, particularly when risk emerges from interactions between individually benign image and question content. To address this issue, we propose A$^2$Safe, a counterfactual evidence-aligned adaptive agent collaboration framework for safe and effective VQA. A$^2$Safe organizes localized visual observations, textual intent, and cross-modal risk relations through a Grounded Safety Evidence Board, making the basis of safety decisions explicit. Counterfactual safety evidence alignment enforces invariance to safety-irrelevant changes while requiring appropriate safety-state and response-mode transitions when risk-critical evidence is minimally altered. The resulting evidence state further supports adaptive collaboration, enabling direct answering when grounded evidence is sufficient and invoking policy critique and response revision when evidence is risky, uncertain, or conflicting. Under complementary safety-critical and general VQA protocols, A$^2$Safe achieves a 95.72 SIUO safety score, reduces the benign refusal rate on MOSSBench to 14.67\%, and maintains an average general VQA score of 78.34 with 27.8\% token overhead. These results support counterfactual evidence-aligned adaptive collaboration for safe and effective multimodal question answering. 

\end{abstract}


\section{Introduction}
\label{sec:intro}

\IEEEPARstart{V}{isual} Question Answering (VQA) has evolved from recognition-oriented prediction into a representative setting for evaluating the perception, reasoning, and instruction-following capabilities of Multimodal Large Language Models (MLLMs)~\cite{zhong5,zhong6,vqa,llava,qwen2vl}. As MLLMs increasingly answer open-ended questions about visual content, VQA requires not only generating correct answers but also determining whether and how they should be answered under contextual safety constraints. Safe and effective VQA is therefore an evidence-sensitive reasoning problem.

Given an image-question pair, a safe and effective VQA system should determine whether the input is benign, context-sensitive, or unsafe and accordingly produce a direct answer, constrained answer, or refusal. Unlike language-only safety, multimodal risk may arise from textual intent, visual evidence, or their interaction. Visual perturbations and heuristic biases can induce reliance on unreliable or superficial multimodal cues~\cite{xu2026etv,lu2026minbias}; typographic attacks and adversarial images may bypass language-side safeguards~\cite{mmsafetybench,figstep}; and individually benign visual and textual content may become unsafe through their composition or situational context~\cite{safeinput,mmsituation}. Effective safe VQA therefore requires not only a correct safety outcome, but also evidence that supports that outcome while avoiding both unsafe compliance and unnecessary refusal.

Recent multimodal safety-alignment methods improve robustness through safety instruction tuning~\cite{safetyalmostnocost,spavl}, preference or reinforcement learning~\cite{saferlhfv}, visual safety prompting~\cite{davsp}, policy-guided reasoning~\cite{teachsafe}, context-dependent reward modeling~\cite{pragmavl}, and rule-governed optimization~\cite{safegrpo}. Despite these advances, correct safety behavior does not necessarily imply evidence-faithful reasoning. A model may associate particular objects, lexical patterns, or visual styles with refusal without identifying the visual-textual relation that actually determines risk~\cite{zhong4,li2026compovis,yuan2025rie}. This limitation is particularly important for context-sensitive inputs: changing a safety-critical entity, action, intent, or relation should alter the safety decision, whereas irrelevant appearance or background changes should not.

\begin{figure}[!t]
	\centering
	\includegraphics[width = 0.8 \linewidth]{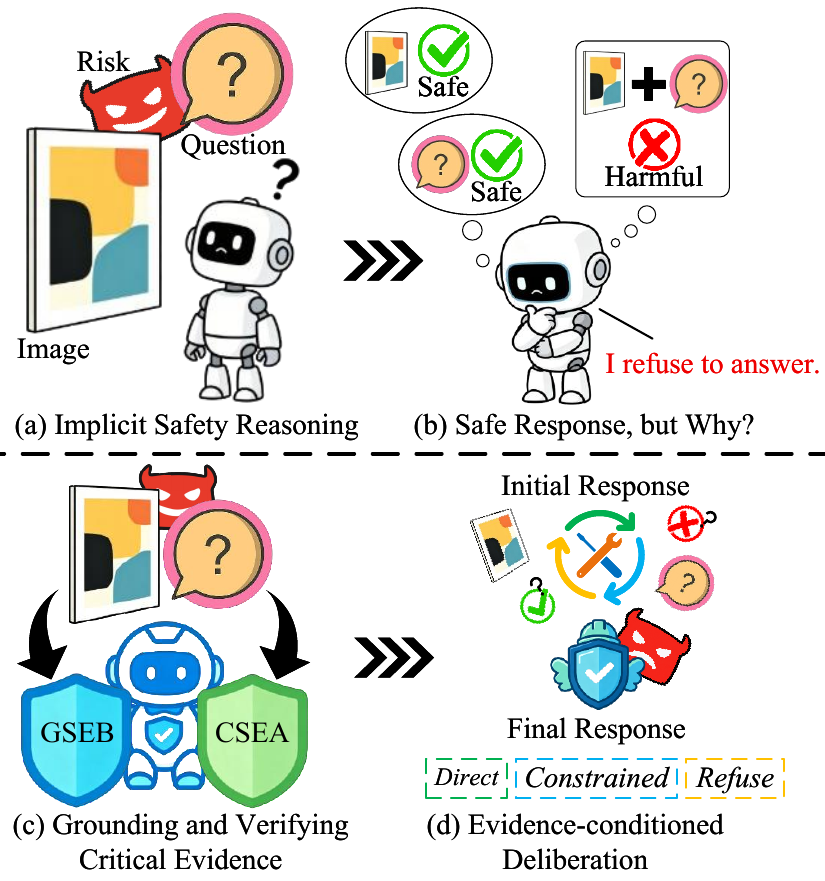}
	\caption{\textbf{Overview of A$^2$Safe.}
    The top (a--b) shows that existing multimodal safety methods may produce safe responses through implicit and unverified safety reasoning. The bottom (c--d) presents A$^2$Safe, which grounds and counterfactually verifies safety-critical evidence and adaptively deliberates over the initial response to obtain an evidence-faithful final response.}
	\label{fig:overview}
\end{figure}

This motivates an evidence-grounded alignment perspective in which localized visual observations, textual intent, and cross-modal relations are explicitly represented, and safety decisions are verified against changes in the evidence that actually determines risk~\cite{xue2025linin,zhong5}. Agent specialization provides a practical way to separate evidence extraction, policy verification, and response generation~\cite{multiagentvqa,alignmentwaltz}. However, collaboration alone is insufficient: without structured grounding and explicit verification, agents may merely exchange unconstrained rationales, while unnecessary deliberation increases inference cost.

To address these limitations, we propose A$^2$Safe, a \textit{Counterfactual Evidence-Aligned Adaptive Agent Collaboration} framework for safe and effective VQA. As shown in \cref{fig:overview}, A$^2$Safe organizes safety-relevant information through a Grounded Safety Evidence Board that contains localized visual evidence, textual intent, cross-modal relations, modality-specific risk, and evidence confidence. An Evidence Agent constructs this structured state, a Response Agent generates an initial answer, and a Policy-Critic Agent verifies candidate responses against grounded evidence and safety policies. This role separation makes agent collaboration an evidence-grounded verification process rather than unconstrained multi-agent discussion.

The core mechanism couples counterfactual safety evidence alignment with evidence-conditioned adaptive collaboration. Safety-preserving transformations enforce invariance to safety-irrelevant changes, whereas safety-flipping transformations require appropriate transitions in the predicted safety state and response mode when critical visual, textual, or relational evidence is minimally altered. By making safety decisions stable to irrelevant variations yet sensitive to risk-critical changes, this alignment discourages reliance on superficial correlations and produces an evidence state for adaptive deliberation. Sufficiently grounded cases follow a direct path, while cases with elevated risk, low confidence, response uncertainty, or inter-agent disagreement invoke policy critique and response revision. Improvement-based collaborative optimization further discourages redundant or harmful revisions, concentrating additional deliberation on cases where verification is most needed.

The contributions of this work are threefold:

\begin{itemize}
	\item We propose an evidence-grounded adaptive agent collaboration framework for safe and effective VQA, using a structured Grounded Safety Evidence Board to expose localized visual observations, textual intent, and cross-modal risk relations underlying safety decisions.

	\item We introduce Counterfactual Safety Evidence Alignment to distinguish risk-critical evidence from superficial correlations. Complementary safety-preserving and safety-flipping interventions enforce invariance to irrelevant variations while maintaining sensitivity to minimal changes that alter the safety state or permissible response mode.

	\item We develop evidence-conditioned adaptive collaboration that allocates policy critique and response revision according to risk, confidence, uncertainty, and inter-agent disagreement, retaining the benefits of collaborative verification without imposing fixed multi-round computation on straightforward inputs.
\end{itemize}

\section{Related Work}
\label{sec:relate}

\subsection{MLLM-Based Visual Question Answering}

Visual Question Answering (VQA) requires models to answer natural-language questions grounded in visual content~\cite{vqa}. By integrating visual encoders with pretrained language models, MLLMs such as LLaVA~\cite{llava} and Qwen2-VL~\cite{qwen2vl} have extended VQA toward open-ended perception, knowledge, and cross-modal reasoning. Recent studies also explore compact visual-semantic cues for task-relevant evidence~\cite{xu2026concise} and increasingly consider consistency and reliability beyond answer accuracy~\cite{xu2026refined}. VLSU further shows that multimodal safety failures may emerge only under joint image-text interpretation~\cite{vlsu}, motivating evidence-sensitive safety reasoning in MLLM-based VQA.

\subsection{Safety Alignment for MLLMs}

Multimodal safety benchmarks expose vulnerabilities beyond language-only safeguards. MM-SafetyBench~\cite{mmsafetybench}, FigStep~\cite{figstep}, SIUO~\cite{safeinput}, and Multimodal Situational Safety~\cite{mmsituation} evaluate malicious, compositional, and context-dependent risks. Corresponding alignment methods include supervised safety tuning~\cite{safetyalmostnocost}, preference alignment~\cite{spavl}, constrained reinforcement learning~\cite{saferlhfv}, policy-guided reasoning~\cite{teachsafe}, visual safety prompting~\cite{davsp}, context-dependent arbitration~\cite{pragmavl}, and rule-governed optimization~\cite{safegrpo}.

Despite these advances, correct safety behavior does not necessarily indicate reliance on the multimodal evidence that determines risk. Safety Mirage shows that safety fine-tuning can exploit superficial correlations~\cite{safetymirage}, while VLSU highlights failures caused by insufficient joint multimodal understanding~\cite{vlsu}. Our work therefore focuses on explicit evidence grounding and counterfactual verification of safety-critical factors.

\subsection{Collaborative Agents for Multimodal Safety}

Collaborative agents provide a practical means of separating evidence extraction, policy verification, and response generation. Multi-Agent VQA coordinates foundation models and visual tools for zero-shot VQA~\cite{multiagentvqa}, while Alignment Waltz jointly trains conversation and feedback agents using adaptive interaction and improvement-based rewards~\cite{alignmentwaltz}. However, existing collaboration does not explicitly verify whether safety decisions are grounded in localized visual, textual, and cross-modal evidence.

A$^2$Safe addresses this limitation by coordinating Evidence, Policy-Critic, and Response Agents through a Grounded Safety Evidence Board. Counterfactual supervision enforces evidence sensitivity, while adaptive routing determines when additional policy deliberation is required. In this way, collaboration becomes a mechanism for grounded and verifiable multimodal safety alignment rather than an end in itself.

\begin{figure}[!t]
	\centering
	\includegraphics[width = 1.0\linewidth]{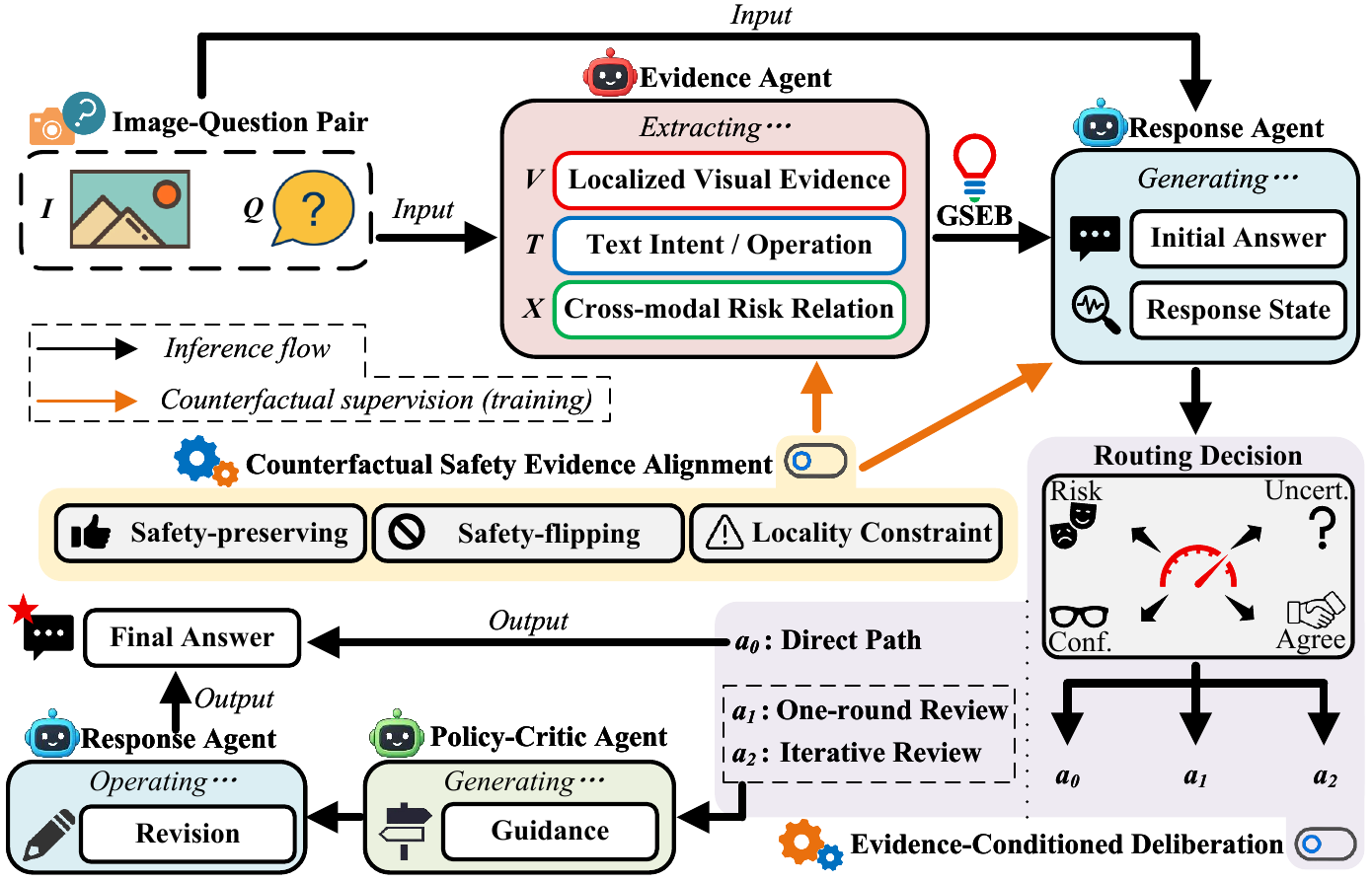}
	\caption{\textbf{Training and inference pipeline of A$^2$Safe.}
    The Evidence Agent constructs the Grounded Safety Evidence Board, while the Response Agent generates an initial response and response state. The evidence-conditioned router determines whether the response is returned directly or refined through Policy-Critic-guided revision. Counterfactual safety evidence alignment and improvement-based collaboration objectives enhance evidence sensitivity and adaptive refinement. Solid and dashed arrows denote inference and training-only paths, respectively.}
	\label{fig:pipeline}
\end{figure}

\section{Proposed Method}
\label{sec:metho}

\cref{fig:pipeline} illustrates the proposed A$^2$Safe framework. We first formulate evidence-grounded safe and effective VQA (\cref{sec:problem}), then introduce the evidence-grounded safety pipeline and Grounded Safety Evidence Board (\cref{sec:framework,sec:evidence}). We subsequently present Counterfactual Safety Evidence Alignment (\cref{sec:counterfactual}), evidence-conditioned adaptive deliberation (\cref{sec:routing}), and improvement-based collaborative optimization (\cref{sec:alignment}). The complete training procedure is provided in the supplementary material.

\subsection{Problem Formulation}
\label{sec:problem}

Let $x=(I,q)$ denote a VQA instance consisting of an image $I$ and a natural-language question or instruction $q$. Conventional VQA predicts an answer according to $p(y\mid I,q)$. Safe and effective VQA additionally requires the model to infer a context-dependent safety state $z$ and select a response mode $d$ from
$\mathcal{D}=\{\texttt{direct},\texttt{constrained},\texttt{refuse}\}$.
Rather than treating safety as response-level refusal prediction, we explicitly condition the decision on multimodal evidence $\mathcal{E}$:
\begin{align}
    (\hat{d},\hat{y})
    = \arg\max_{d\in\mathcal{D}, y}
    p(d,y\mid I,q,\mathcal{E}),
    \label{eq:task}
\end{align}
where $\mathcal{E}$ contains visual observations, textual intent, and image-text relations supporting the safety judgment. The three response modes correspond to benign inputs that permit direct answers, context-sensitive inputs that permit bounded assistance, and unsafe requests that require refusal. This formulation makes both the safety outcome and its supporting evidence explicit.

A$^2$Safe implements this formulation using three role-specialized agents
$\mathcal{A}_{k}$, where $k\in\{E,P,R\}$ denotes the Evidence, Policy-Critic, and Response roles. For efficiency, the agents share a pretrained MLLM backbone $M_{\theta}$ and use independent lightweight role adapters $\phi_k$:
\begin{align}
    \mathcal{A}_{k}(\cdot)
    = M_{\theta,\phi_k}(\cdot),
    \quad k\in\{E,P,R\}.
    \label{eq:agents}
\end{align}
The backbone parameters $\theta$ remain frozen, while
$\{\phi_E,\phi_P,\phi_R\}$ and the router parameters are jointly optimized. This role decomposition separates evidence extraction, policy verification, and response generation while retaining a shared multimodal representation space.

\subsection{Evidence-Grounded Safety Pipeline}
\label{sec:framework}

Given $x$, the Evidence Agent first extracts safety-relevant multimodal observations and records them in a structured evidence board $\mathcal{B}$. Conditioned on $\mathcal{B}$, the Response Agent generates an initial answer $y^{(0)}$ and a safety distribution
$p_R(z\mid I,q,\mathcal{B},y^{(0)})$ over benign, context-sensitive, and unsafe states. An evidence-conditioned router then selects
$a\in\mathcal{A}=\{a_0,a_1,a_2\}$, corresponding to direct answering, one-round review, and iterative review, respectively. Routing is thus performed after both grounded evidence and an actual candidate response are available.

For a routed review round $t$, the Policy-Critic Agent reads the input, evidence board, and current response and produces structured feedback $f^{(t)}$ containing the recommended response mode, cited evidence, relevant policy, and revision instruction. The Response Agent then revises the answer as:
\begin{align}
    f^{(t)}
    &= \mathcal{A}_{P}
    \left(I,q,\mathcal{B},y^{(t)}\right), \\
    y^{(t+1)}
    &= \mathcal{A}_{R}
    \left(I,q,\mathcal{B},y^{(t)},f^{(t)}\right).
    \label{eq:interaction}
\end{align}
During training, deliberation terminates when the critic accepts the response, the measured improvement falls below $\epsilon$, or the maximum number of rounds $T_{\max}$ is reached. At inference, reward computation is removed, and termination depends only on critic acceptance and $T_{\max}$. Collaboration therefore acts as selective policy deliberation rather than a mandatory stage for every input.

\subsection{Grounded Safety Evidence Board}
\label{sec:evidence}

Multimodal risk may originate from the image, the question, or their composition. The Evidence Agent therefore constructs:
\begin{align}
    (\mathcal{B},p_E)
    &= \mathcal{A}_{E}(I,q), \\
    \mathcal{B}
    &= \{\mathcal{V},\mathcal{T},\mathcal{X},
        r_v,r_t,r_{vt},c_E\},
    \label{eq:board}
\end{align}
where $\mathcal{V}$ contains localized objects, attributes, actions, and OCR tokens, while $\mathcal{T}$ represents textual intent and the requested operation. The relation set $\mathcal{X}$ explicitly represents cross-modal safety evidence. Each relation is encoded as
$e_j^{x}=(v_j,t_j,\rho_j,\gamma_j)$, where $v_j$ denotes grounded visual evidence, $t_j$ the associated question span or intent, $\rho_j$ their semantic relation, and $\gamma_j$ its relevance to the safety decision. The scalars $r_v$, $r_t$, and $r_{vt}\in[0,1]$ denote visual, textual, and compositional risk, respectively, while $c_E$ denotes confidence in the grounded evidence.

The categorical distribution $p_E$ is defined over
$\{\texttt{benign},\texttt{context-sensitive},\texttt{unsafe}\}$.
In implementation, $p_E$ and the Response Agent distribution $p_R$ are obtained from lightweight classification heads applied to pooled hidden representations of their respective role adapters, followed by softmax. Similarly, the Policy-Critic distribution $p_P$ is produced by a classification head over
$\{\texttt{direct},\texttt{constrained},\texttt{refuse}\}$.
These distributions are explicit discriminative predictions rather than probabilities derived from designated generation tokens. Each evidence item is associated with a visual region or textual span, requiring subsequent decisions to refer to verifiable observations rather than unrestricted rationales.

For role-specific supervision, let $\bar{z}$, $\bar{d}$, $\bar{y}$, and $\bar{\mathcal{B}}$ denote the target safety state, response mode, reference answer, and evidence board. The board objective is:
\begin{align}
\mathcal{L}_{\mathrm{board}}
= & \lambda_v\mathcal{L}_{\mathrm{ent}}
+\lambda_t\mathcal{L}_{\mathrm{intent}}
+\lambda_x\mathcal{L}_{\mathrm{rel}} \nonumber\\
&+\lambda_r\mathcal{L}_{\mathrm{region}}
+\lambda_{\rm conf}\mathcal{L}_{\mathrm{conf}},
\label{eq:boardloss}
\end{align}
where the terms supervise visual entities, textual intent, cross-modal relations, cited regions or spans, and calibrated evidence confidence. The role-specialization objective is:
\begin{align}
    \mathcal{L}_{\mathrm{role}}
    = & \mathcal{L}_{\mathrm{CE}}(p_E,\bar{z})
    + \lambda_R\mathcal{L}_{\mathrm{CE}}(p_R,\bar{z})
    + \lambda_d\mathcal{L}_{\mathrm{CE}}(p_P,\bar{d}) \nonumber\\
    &- \lambda_y\log \pi_R(\bar{y}\mid I,q,\mathcal{B})
    + \lambda_b\mathcal{L}_{\mathrm{board}}
      (\mathcal{B},\bar{\mathcal{B}}).
    \label{eq:roleloss}
\end{align}
We construct $\bar{\mathcal{B}}$ using training-only safety labels and available region/OCR annotations, with a fixed teacher proposing missing fields. A proposal is retained only when region grounding, OCR consistency, and policy-label checks agree, limiting unrestricted rationale distillation and preventing test leakage.

\subsection{Counterfactual Safety Evidence Alignment}
\label{sec:counterfactual}

Evidence grounding alone does not guarantee that a correct safety decision depends on the grounded safety-critical evidence; models may still exploit spurious objects, lexical triggers, or dataset-specific visual patterns. We therefore verify evidence dependence through two complementary counterfactual interventions.

A safety-preserving transformation $T^{+}$ produces
$x^{+}=T^{+}(x)$ while retaining the original safety state and response mode. Examples include benign paraphrasing, irrelevant background changes, and appearance variations that preserve the safety-critical entity, action, and image-text relation. We enforce invariance through:
\begin{align}
    \mathcal{L}_{\mathrm{pre}}
    = & D_{\mathrm{SKL}} \left(p_E(x),p_E(x^{+})\right)
       +D_{\mathrm{SKL}} \left(p_P(x),p_P(x^{+})\right) \nonumber\\
       &+\lambda_x
       \left\|h_{\mathcal{X}}(x)-h_{\mathcal{X}}(x^{+})\right\|_1,
    \label{eq:preserve}
\end{align}
where $D_{\mathrm{SKL}}$ denotes symmetric KL divergence and
$h_{\mathcal{X}}(\cdot)$ is a fixed-dimensional pooled representation of the relation set $\mathcal{X}$.

Conversely, a safety-flipping transformation $T^{-}$ minimally modifies safety-critical evidence, yielding $x^{-}=T^{-}(x)$ with $z^{-}\neq z$. The intervention may alter a critical action, requested intent, or image-question relation while preserving unrelated content. Because the safety state contains three ordered risk levels, we define the ordinal risk expectation:
\begin{equation}
    s(x)=\sum_{k=1}^{3}\omega_k p_E(z=z_k\mid x),
    \quad
    \bm{\omega}=[0,0.5,1],
    \label{eq:ordinalrisk}
\end{equation}
where
$z_k\in\{\texttt{benign},\texttt{context-sensitive},\texttt{unsafe}\}$.
This formulation captures increasing safety risk while preserving the full three-state prediction. Note that we use $\bm{\omega}=[0,0.5,1]$ as a simple normalized ordinal encoding of benign, context-sensitive, and unsafe states, respectively, so that $s(x)\in[0,1]$ represents the expected risk level under the evidence-agent prediction.

Counterfactual sensitivity is then enforced by:
\begin{align}
    \mathcal{L}_{\mathrm{flip}}
    = & \left[
    m-\rho(x,x^{-})
    \left(s(x)-s(x^{-})\right)
    \right]_{+} \nonumber\\
    &+\lambda_z\mathcal{L}_{\mathrm{CE}}
      \left(p_E(x^{-}),z^{-}\right)
    +\lambda_f\mathcal{L}_{\mathrm{CE}}
      \left(p_P(x^{-}),d^{-}\right),
    \label{eq:flip}
\end{align}
where $m$ is a margin and
$\rho(x,x^{-})=1$ if the target state of $x$ is riskier than that of $x^{-}$ and $-1$ otherwise. The first term enforces the correct direction of the risk transition, while the remaining terms supervise the counterfactual safety state and corresponding response mode.

To ensure that the transition is attributable to the intended intervention rather than unintended semantic changes, we additionally preserve representations of non-critical evidence:
\begin{align}
    \mathcal{L}_{\mathrm{loc}}
    =
    \left\|
    h_{\neg c}(x)-h_{\neg c}(x^{-})
    \right\|_1,
    \label{eq:locality}
\end{align}
where $h_{\neg c}$ pools evidence outside the intervened safety-critical region or span. Generated pairs are further filtered by semantic-validity and evidence-locality checks.

The complete counterfactual objective is:
\begin{align}
    \mathcal{L}_{\mathrm{CSEA}}
    = & \mathcal{L}_{\mathrm{board}}(x^{+})
    +\mathcal{L}_{\mathrm{board}}(x^{-})
    +\lambda_{+}\mathcal{L}_{\mathrm{pre}} \nonumber\\
    &+\lambda_{-}\mathcal{L}_{\mathrm{flip}}
    +\lambda_{\mathrm{loc}}\mathcal{L}_{\mathrm{loc}}.
    \label{eq:csea}
\end{align}
CSEA therefore jointly encourages invariance to safety-irrelevant variations, sensitivity to safety-critical interventions, and locality of the induced change.

\subsection{Evidence-Conditioned Adaptive Deliberation}
\label{sec:routing}

Invoking policy critique for every sample increases computation and may unnecessarily modify already-correct benign responses. The router $g_{\psi}$ therefore estimates whether the current evidence and response are sufficiently reliable for direct answering. It jointly considers compositional risk $r_{vt}$, evidence uncertainty $1-c_E$, normalized response entropy $\mathcal{H}(p_R)$, and Jensen-Shannon disagreement between the Evidence and Response Agents:
\begin{align}
    u(x)
    &= g_{\psi} \left(
    r_{vt},1-c_E,\mathcal{H}(p_R),
    D_{\mathrm{JS}}(p_E,p_R)
    \right), \\
    a(x)
    &=
    \begin{cases}
       a_0, & u(x)<\tau_1,\\
       a_1, & \tau_1\leq u(x)<\tau_2,\\
       a_2, & u(x)\geq\tau_2.
    \end{cases}
    \label{eq:router}
\end{align}
The routing score thus reflects not only predicted risk, but also evidence confidence, response uncertainty, and consistency between evidence and response judgments.

During training, a response is valid when its response mode matches the target policy, satisfies the safety constraint, and, for answerable benign or context-sensitive inputs, exceeds a validation threshold for correctness. The target route is the least costly action that produces a valid response, with ties broken by the utility in \cref{eq:utility}. The router is optimized with route supervision and an execution-cost regularizer. At inference, $a_0$ returns $y^{(0)}$, $a_1$ invokes one critique-revision round, and $a_2$ permits up to $T_{\max}$ rounds with early stopping.

\subsection{Improvement-Based Collaborative Optimization}
\label{sec:alignment}

The collaborative agents are optimized so that additional deliberation is beneficial rather than merely longer. Following joint agent optimization~\cite{alignmentwaltz}, we evaluate each response using a rule-governed utility adapted from multimodal safety rewards~\cite{safegrpo}:
\begin{align}
    J(y^{(t)}\mid x)
    = &
    w_c R_{\mathrm{cor}}^{(t)}
    + w_s R_{\mathrm{safe}}^{(t)}
    + w_h R_{\mathrm{help}}^{(t)} \nonumber\\
    &-\lambda_o R_{\mathrm{over}}^{(t)}
    -\lambda_g R_{\mathrm{ung}}^{(t)}
    -\lambda_e C^{(t)},
    \label{eq:utility}
\end{align}
where $R_{\mathrm{cor}}$ evaluates correctness for answerable benign or context-sensitive questions, $R_{\mathrm{safe}}$ measures consistency among the target safety state, response mode, and answer behavior, and $R_{\mathrm{help}}$ measures permissible information retained in constrained responses or refusals. $R_{\mathrm{over}}$ penalizes unnecessary restriction on benign inputs, $R_{\mathrm{ung}}$ penalizes claims unsupported by the evidence board, and $C^{(t)}$ measures computation through agent calls and generated tokens. We use deterministic checks where possible and a fixed semantic judge for correctness, helpfulness, and grounding.

For round $t$, the improvement reward is:
\begin{align}
    R_{\mathrm{imp}}^{(t)}
    = J(y^{(t+1)}\mid x)-J(y^{(t)}\mid x).
    \label{eq:improvement}
\end{align}
Feedback therefore receives positive reward only when it measurably improves the response. The Policy-Critic and Response policies are jointly optimized with KL regularization:
\begin{align}
\mathcal{L}_{\mathrm{collab}}
= &-\mathbb{E}_{x,t}\Big[
R_{\mathrm{imp}}^{(t)}
\Big(
\log\pi_P(f^{(t)}\mid s_t) \nonumber\\
&
+\eta\log\pi_R
\left(y^{(t+1)}\mid s_t,f^{(t)}\right)
\Big)\Big] \nonumber\\
&+\beta_P D_{\mathrm{KL}}
\left(\pi_P\|\pi_P^{\mathrm{ref}}\right)
+\beta_R D_{\mathrm{KL}}
\left(\pi_R\|\pi_R^{\mathrm{ref}}\right), \\
\mathcal{L}_{\mathrm{A}^{2}\mathrm{Safe}}
= &\mathcal{L}_{\mathrm{role}}
+\lambda_c\mathcal{L}_{\mathrm{CSEA}}
+\lambda_a\mathcal{L}_{\mathrm{collab}}
+\lambda_r\mathcal{L}_{\mathrm{route}},
\label{eq:total}
\end{align}
where $s_t=(I,q,\mathcal{B},y^{(t)})$ denotes the collaborative state, and $\mathcal{L}_{\mathrm{route}}$ contains route-supervision and execution-cost terms. In Stage I, we set $\lambda_a=\lambda_r=0$ and optimize role specialization and counterfactual evidence alignment. In Stage II, the collaborative and routing objectives are activated while the Stage-I objectives are retained as regularizers. The shared backbone remains frozen throughout; only the role adapters and router are updated.

During training, sequential oracle rollouts identify the least-cost valid route. At inference, counterfactual branches, oracle rollouts, and reward computation are removed, leaving only the Grounded Safety Evidence Board, initial response, and evidence-conditioned router.

\section{Experimental Results}
\label{sec:experi}

We evaluate A$^2$Safe from three complementary perspectives: overall safety and VQA capability, evidence-grounded and counterfactual safety behavior, and adaptive deliberation efficiency. We first describe the evaluation setup, then compare A$^2$Safe with representative safety-alignment and collaborative methods, followed by mechanism-level ablation, counterfactual, and routing analyses.

\subsection{Evaluation Setup}
\label{sec:setup}

\subsubsection{Evaluation Protocol and Metrics}
\label{sec:metrics}

The evaluation covers explicit jailbreak resistance, contextual multimodal safety, over-refusal, general VQA capability, evidence-grounded safety reasoning, counterfactual behavior, and inference efficiency. Following recent multimodal safety studies~\cite{safegrpo,thinkinsafety}, we report Safety Score (SS, $\uparrow$) on FigStep, VLGuard, and MM-SafetyBench, with Avg.~SS denoting their mean. SIUO and MSSBench evaluate cross-modal and situational safety~\cite{safeinput,mmsituation}, while the MOSSBench Refusal Rate (RR, $\downarrow$) measures over-refusal on benign but visually sensitive queries~\cite{mossbench}.

General VQA capability is evaluated on ScienceQA~\cite{scienceqa}, IconQA~\cite{iconqa}, MathVista~\cite{mathvista}, MM-Vet~\cite{mmvet}, and POPE~\cite{pope}. We additionally follow the Pragma-VL protocol~\cite{pragmavl} for safety-helpfulness arbitration and use EviSafeBench~\cite{evisafe} to assess natural safety accuracy, evidence grounding, diagnostic consistency, and counterfactual transition success.

Preservation Consistency (PC, $\uparrow$) measures invariance under safety-preserving changes, while Flip Sensitivity (FS, $\uparrow$) measures correct safety-state and response-mode transitions after safety-critical interventions. For adaptive deliberation, we further report route accuracy, normalized route regret, average critique-revision rounds, token overhead, and relative latency.

\subsubsection{Datasets and Tasks}
\label{sec:datasets}

\paragraph{Alignment Data}
We train exclusively on the training partitions of VLGuard~\cite{safetyalmostnocost} and a 30K subset of SPA-VL~\cite{spavl}. Potential overlap with evaluation benchmarks is removed using perceptual image hashing, normalized question similarity, and manual inspection. For eligible samples, safety-preserving and safety-flipping counterparts are constructed to form the counterfactual training set $\mathcal{D}_{\rm cf}$. Detailed annotation and filtering procedures are provided in the supplementary material.

\paragraph{Safety Evaluation}
We use FigStep~\cite{figstep}, MM-SafetyBench~\cite{mmsafetybench}, VLGuard~\cite{safetyalmostnocost}, SIUO~\cite{safeinput}, MSSBench~\cite{mmsituation}, and MOSSBench~\cite{mossbench} for outcome-level evaluation, covering explicit attacks, cross-modal composition, situational safety, and over-refusal. We use EviSafeBench~\cite{evisafe} specifically for evidence-grounded diagnosis.

\paragraph{General Capability Evaluation}
General VQA capability is evaluated on ScienceQA~\cite{scienceqa}, IconQA~\cite{iconqa}, MathVista~\cite{mathvista}, MM-Vet~\cite{mmvet}, and POPE~\cite{pope}. No samples from these benchmarks are used for safety alignment.

\subsubsection{Implementation Details}
\label{sec:implementation}

We use Qwen3-VL-4B-Thinking as the default backbone~\cite{qwen3vl,safegrpo}. The three agents share a frozen backbone and use independent LoRA adapters. The evidence-conditioned router takes compositional risk, evidence uncertainty $1-c_E$, response entropy, and Evidence-Response disagreement as input. Training follows the two-stage procedure in \cref{sec:alignment}: role specialization and CSEA are optimized first, followed by collaborative and routing optimization. For the complementary Pragma-VL protocol, we use Qwen2.5-VL-7B under the same framework. Detailed hyperparameters, hardware settings, evaluation controls, and annotation procedures are provided in the supplementary material.

\subsubsection{Baselines and Comparison Protocol}

We compare A$^2$Safe with representative supervised, preference-based, inference-time, reasoning-oriented, and collaborative safety methods, including VLGuard~\cite{safetyalmostnocost}, SPA-VL~\cite{spavl}, Safe RLHF-V~\cite{saferlhfv}, ECSO~\cite{ecso}, Think in Safety~\cite{thinkinsafety}, SafeGRPO~\cite{safegrpo}, Pragma-VL~\cite{pragmavl}, Multi-Agent VQA~\cite{multiagentvqa}, and Alignment Waltz~\cite{alignmentwaltz}. Controlled reproductions use matched backbones, training data, decoding settings, evaluators, and interaction budgets. Details of multimodal baseline adaptations are provided in the supplementary material.

\begin{table*}[!t]
\centering
\caption{\textbf{Safety performance under the Qwen3-VL-4B-Thinking protocol.}
SS denotes Safety Score, and RR denotes the benign refusal rate. Best results are shown in \textbf{bold}, and second-best results are \underline{underlined}.}
\label{tab:safety-main}

\setlength{\tabcolsep}{7pt}

\begin{tabular}{ll|cccccc}
\toprule[1.1pt]
\textbf{Method} &
\textbf{Venue} &
\textbf{FigStep SS $\uparrow$} &
\textbf{VLGuard SS $\uparrow$} &
\textbf{MM-Safety SS $\uparrow$} &
\textbf{Avg. SS $\uparrow$} &
\textbf{SIUO $\uparrow$} &
\textbf{MOSS RR $\downarrow$} \\
\midrule

Qwen3-VL-4B-Thinking
& Tech. Rep.'25
& 88.08
& 93.24
& 90.48
& 90.60
& 89.88
& 27.00 \\

+ VLGuard~\cite{safetyalmostnocost}
& ICML'24
& 96.32
& 97.96
& 96.29
& 96.86
& 90.41
& 98.33 \\

+ ECSO~\cite{ecso}
& ECCV'24
& 90.30
& 95.97
& 97.17
& 94.48
& 89.76
& 29.00 \\

+ Think in Safety~\cite{thinkinsafety}
& EMNLP'25
& 98.40
& 98.05
& 97.19
& 97.88
& 91.31
& 68.67 \\

+ SafeGRPO~\cite{safegrpo}
& CVPR'26
& \underline{99.60}
& 98.64
& \underline{99.38}
& \underline{99.21}
& \underline{93.85}
& 24.33 \\

\rowcolor{gray!15}
+ A$^2$Safe
& Ours
& \textbf{99.68}
& \textbf{99.01}
& \textbf{99.40}
& \textbf{99.36}
& \textbf{95.72}
& \textbf{14.67} \\

\bottomrule[1.1pt]
\end{tabular}
\end{table*}

\subsection{Main Results}
\label{sec:overall}

\subsubsection{Explicit and Contextual Safety}

\cref{tab:safety-main} shows that explicit-attack safety is already near saturation: SafeGRPO achieves 99.21 Avg.~SS, while A$^2$Safe reaches 99.36. Larger differences emerge in context-dependent behavior. A$^2$Safe improves SIUO from 93.85 to 95.72 and reduces MOSSBench RR from 24.33\% to 14.67\%, whereas VLGuard and Think in Safety reject 98.33\% and 68.67\% of benign queries, respectively. These results indicate that evidence-sensitive alignment primarily improves contextual discrimination rather than uniformly increasing refusal.

\begin{figure*}[!t]
	\centering
	\includegraphics[width=\linewidth]{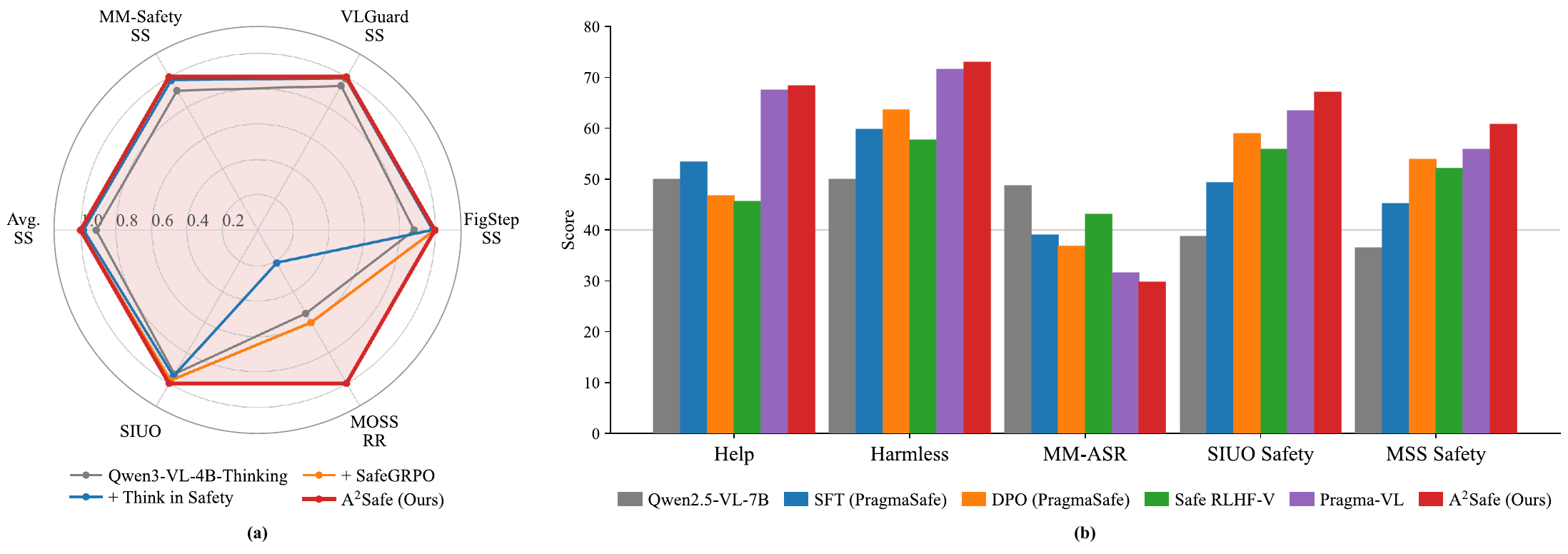}
	\caption{\textbf{Safety and safety-helpfulness comparison.}
	(a) Safety performance under the Qwen3-VL-4B-Thinking protocol across explicit jailbreak defense, contextual safety, and benign over-refusal; MOSS RR is direction-normalized so that larger radial values indicate better performance.
	(b) Safety-helpfulness comparison under the Qwen2.5-VL-7B protocol, jointly reporting Help, Harmless, MM-Safety ASR, SIUO Safety, and MSS Safety.}
	\label{fig:main-effectiveness}
\end{figure*}

\cref{fig:main-effectiveness}(a) further illustrates the contrast between explicit-attack safety and benign over-refusal. A$^2$Safe maintains strong safety while reducing unnecessary refusal on visually sensitive benign inputs, consistent with its evidence-grounded response modes.

\begin{table*}[!t]
\centering
\caption{\textbf{Safety-helpfulness comparison under the Qwen2.5-VL-7B protocol.}
Baseline results follow the unified reproductions reported by Pragma-VL~\cite{pragmavl}. 
$^{\dag}$ denotes reproduction under the matched Qwen2.5-VL-7B evaluation protocol. 
Best results are shown in \textbf{bold}, and second-best results are \underline{underlined}.}
\label{tab:pragma-protocol}

\setlength{\tabcolsep}{11pt}

\begin{tabular}{ll|ccccc}
\toprule[1.1pt]
\textbf{Method} &
\textbf{Venue} &
\textbf{Help $\uparrow$} &
\textbf{Harmless $\uparrow$} &
\textbf{MM-Safety ASR $\downarrow$} &
\textbf{SIUO Safety $\uparrow$} &
\textbf{MSS Safety $\uparrow$} \\
\midrule

Qwen2.5-VL-7B
& Tech. Rep.'25
& 50.00
& 50.00
& 48.75
& 38.78
& 36.53 \\

SFT (PragmaSafe)~\cite{pragmavl}
& Reproduced
& 53.36
& 59.91
& 39.07
& 49.39
& 45.28 \\

DPO (PragmaSafe)~\cite{pragmavl}
& Reproduced
& 46.75
& 63.60
& 36.79
& 59.03
& 53.96 \\

Safe RLHF-V~\cite{saferlhfv}
& NeurIPS'25$^{\dag}$
& 45.70
& 57.73
& 43.20
& 55.90
& 52.20 \\

Pragma-VL~\cite{pragmavl}
& ICLR'26
& \underline{67.52}
& \underline{71.61}
& \underline{31.66}
& \underline{63.47}
& \underline{55.89} \\

\rowcolor{gray!15}
A$^2$Safe
& Ours
& \textbf{68.40}
& \textbf{73.10}
& \textbf{29.80}
& \textbf{67.20}
& \textbf{60.80} \\

\bottomrule[1.1pt]
\end{tabular}
\end{table*}

\subsubsection{Safety-Helpfulness Arbitration}

\cref{tab:pragma-protocol} shows that SFT and DPO improve harmlessness at the cost of useful information, while Pragma-VL reaches 67.52 Help and 71.61 Harmless. A$^2$Safe further improves these scores to 68.40 and 73.10 and achieves larger gains on SIUO and MSSBench, indicating improved arbitration between safety and useful answering in context-dependent multimodal settings. \cref{fig:main-effectiveness}(b) visualizes this trade-off.

\begin{table*}[!t]
\centering
\caption{\textbf{General multimodal capability under the Qwen3-VL-4B-Thinking protocol.}
Published baseline results are taken from SafeGRPO~\cite{safegrpo}. 
Best results are shown in \textbf{bold}, and second-best results are \underline{underlined}.}
\label{tab:utility}

\setlength{\tabcolsep}{10pt}

\begin{tabular}{ll|cccccc}
\toprule[1.1pt]
\textbf{Method} &
\textbf{Venue} &
\textbf{ScienceQA $\uparrow$} &
\textbf{IconQA $\uparrow$} &
\textbf{MathVista $\uparrow$} &
\textbf{MM-Vet $\uparrow$} &
\textbf{POPE $\uparrow$} &
\textbf{Avg. $\uparrow$} \\
\midrule

Qwen3-VL-4B-Thinking
& Tech. Rep.'25
& 85.92
& 83.60
& 60.70
& 63.44
& \underline{87.70}
& 76.27 \\

+ VLGuard~\cite{safetyalmostnocost}
& ICML'24
& 78.73
& 70.30
& 42.20
& 36.29
& 71.10
& 59.72 \\

+ ECSO~\cite{ecso}
& ECCV'24
& 85.92
& 84.00
& 60.70
& 63.44
& \textbf{87.80}
& 76.37 \\

+ Think in Safety~\cite{thinkinsafety}
& EMNLP'25
& 76.05
& 78.20
& 53.00
& 52.24
& 34.20
& 58.74 \\

+ SafeGRPO~\cite{safegrpo}
& CVPR'26
& \underline{87.75}
& \underline{86.20}
& \underline{64.80}
& \underline{64.36}
& 87.40
& \underline{78.10} \\

\rowcolor{gray!15}
+ A$^2$Safe
& Ours
& \textbf{87.90}
& \textbf{86.40}
& \textbf{65.10}
& \textbf{64.70}
& 87.60
& \textbf{78.34} \\

\bottomrule[1.1pt]
\end{tabular}
\end{table*}

\begin{figure}[!t]
	\centering
	\includegraphics[width=0.8\linewidth]{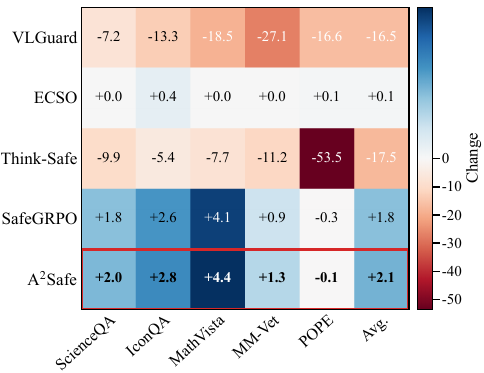}
	\caption{\textbf{General multimodal capability relative to the Qwen3-VL-4B-Thinking backbone.}
	Each cell reports the absolute score change on ScienceQA, IconQA, MathVista, MM-Vet, POPE, and their average after safety alignment. Positive values indicate capability gains over the backbone, whereas negative values indicate degradation.}
	\label{fig:general}
\end{figure}

\subsubsection{General VQA Capability}

\cref{tab:utility} shows that A$^2$Safe achieves 78.34 Avg.~VQA, 2.07 points above the original backbone and 0.24 points above SafeGRPO, while maintaining nearly unchanged POPE performance. This suggests that evidence-grounded safety alignment preserves general multimodal capability. As shown in \cref{fig:general}, the improvement is distributed across multiple benchmarks rather than being driven by a single task.

\begin{table*}[!t]
\centering
\caption{\textbf{Evidence-grounded safety evaluation on EviSafeBench.}
Nat.~Acc. denotes natural safety-decision accuracy, Evid.~Ground. measures identification of safety-critical multimodal evidence, Diag.~Cons. measures consistency between evidence and safety decisions, and CF Trans. measures correct response transitions under targeted counterfactual interventions. Best results are shown in \textbf{bold}, and second-best results are \underline{underlined}.}
\label{tab:evisafe}

\setlength{\tabcolsep}{12pt}

\begin{tabular}{ll|cccc}
\toprule[1.1pt]
\textbf{Method} &
\textbf{Venue} &
\textbf{Nat. Acc. $\uparrow$} &
\textbf{Evid. Ground. $\uparrow$} &
\textbf{Diag. Cons. $\uparrow$} &
\textbf{CF Trans. $\uparrow$} \\
\midrule

Qwen3-VL-4B-Thinking
& Tech. Rep.'25
& 51.8
& 35.6
& 27.9
& 55.1 \\

SafeGRPO~\cite{safegrpo}
& CVPR'26
& 67.4
& 49.8
& 39.6
& 69.2 \\

Pragma-VL~\cite{pragmavl}
& ICLR'26
& 69.1
& 52.7
& 42.3
& 71.5 \\

A$^2$Safe \textit{w/o} CSEA
& Ours
& \underline{71.8}
& \underline{61.9}
& \underline{48.7}
& 67.8 \\

\rowcolor{gray!15}
A$^2$Safe
& Ours
& \textbf{78.6}
& \textbf{72.4}
& \textbf{64.1}
& \textbf{83.7} \\

\bottomrule[1.1pt]
\end{tabular}
\end{table*}

\subsubsection{Evidence-Grounded Safety Evaluation}

\cref{tab:evisafe} examines whether safety decisions are supported by the multimodal evidence that determines risk. The vanilla backbone performs substantially worse on evidence grounding and diagnostic consistency than on natural safety decisions, revealing a gap between safe-looking outcomes and evidence-faithful reasoning.

A$^2$Safe substantially improves evidence grounding, diagnostic consistency, and counterfactual transition performance. Removing CSEA primarily degrades counterfactual transitions, indicating complementary roles for GSEB and CSEA: GSEB identifies candidate safety evidence, while CSEA verifies whether the resulting decision is sensitive to that evidence.


\begin{table*}[!t]
\centering
\caption{\textbf{Comparison with collaborative-agent baselines under the unified Qwen3-VL-4B-Thinking protocol.}
All reproduced methods use matched training data, decoding settings, evaluators, and interaction budgets. Avg.~SS is averaged over FigStep, VLGuard, and MM-SafetyBench. Token $\Delta$ and latency are measured relative to single-pass Qwen3-VL inference. $^{\dagger}$ denotes a multimodal safety adaptation of the original method. Best results are shown in \textbf{bold}, and second-best results are \underline{underlined}.}
\label{tab:agent-comparison}

\setlength{\tabcolsep}{5pt}

\begin{tabular}{ll|ccccccc}
\toprule[1.1pt]
\textbf{Method} &
\textbf{Venue} &
\textbf{Avg. SS $\uparrow$} &
\textbf{SIUO $\uparrow$} &
\textbf{MSS Safety $\uparrow$} &
\textbf{MOSS RR $\downarrow$} &
\textbf{Avg. VQA $\uparrow$} &
\textbf{Token $\Delta$ $\downarrow$} &
\textbf{Latency $\downarrow$} \\
\midrule

Qwen3-VL-4B-Thinking
& Tech. Rep.'25
& 90.60
& 89.88
& 46.72
& 27.00
& 76.27
& \textbf{+0.0\%}
& \textbf{$1.00\times$} \\

SafeGRPO~\cite{safegrpo}
& CVPR'26
& \underline{99.21}
& 93.85
& 50.73
& 24.33
& \underline{78.10}
& \underline{+8.9\%}
& \underline{$1.09\times$} \\

Multi-Agent VQA-S$^{\dagger}$~\cite{multiagentvqa}
& arXiv'24
& 96.74
& 92.31
& 52.18
& 23.67
& 77.15
& +69.4\%
& $1.72\times$ \\

Alignment Waltz-VL$^{\dagger}$~\cite{alignmentwaltz}
& ICLR'26
& 98.63
& \underline{94.28}
& \underline{55.42}
& \underline{18.33}
& 77.68
& +84.7\%
& $1.84\times$ \\

\rowcolor{gray!15}
A$^2$Safe
& Ours
& \textbf{99.36}
& \textbf{95.72}
& \textbf{58.90}
& \textbf{14.67}
& \textbf{78.34}
& +27.8\%
& $1.31\times$ \\

\bottomrule[1.1pt]
\end{tabular}
\end{table*}

\subsubsection{Controlled Comparison with Collaborative Agents}
\label{sec:agent-comparison}

\cref{tab:agent-comparison} shows that collaborative baselines improve contextual safety but incur substantial generation overhead. Alignment Waltz-VL reaches 94.28 SIUO and 18.33\% MOSS RR, whereas A$^2$Safe improves SIUO to 95.72 and MSS Safety to 58.90, reduces MOSS RR to 14.67\%, and limits token overhead to 27.8\%.

\begin{figure}[!t]
	\centering
	\includegraphics[width=0.8\linewidth]{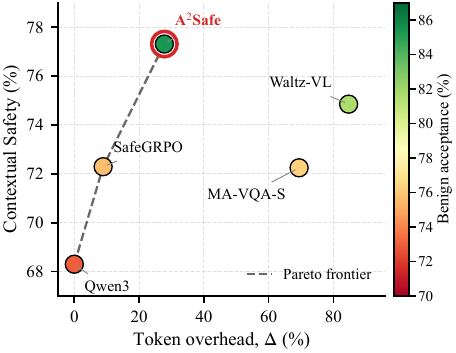}
	\caption{\textbf{Safety--efficiency comparison with collaborative baselines.}
    The horizontal axis reports token overhead relative to single-pass Qwen3-VL inference, while the vertical axis reports contextual safety, defined as the average safety score on SIUO and MSSBench. Marker color denotes benign acceptance rate, and the dashed line indicates the Pareto frontier across compared methods.}
	\label{fig:collaborative}
\end{figure}

\begin{figure*}[!t]
	\centering
	\includegraphics[width=0.9\linewidth]{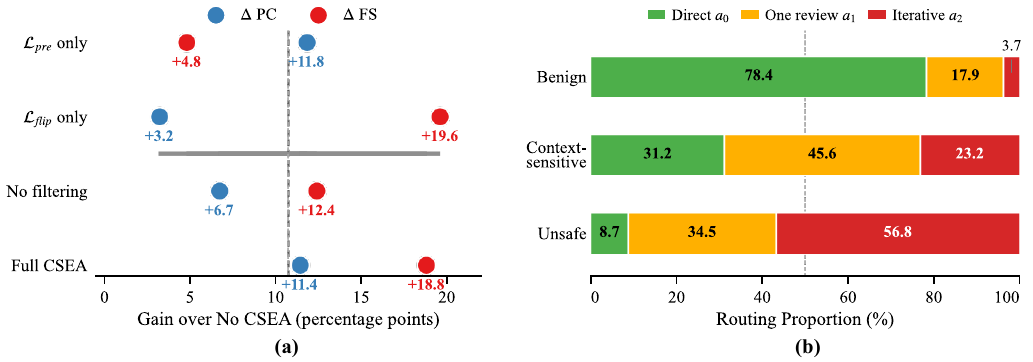}
	\caption{\textbf{Counterfactual evidence verification and adaptive deliberation.}
	(a) Gains in preservation consistency (PC) and flip sensitivity (FS) over the \textit{w/o} CSEA baseline, showing the complementary effects of safety-preserving and safety-flipping supervision as well as the impact of validity filtering.
	(b) Routing distribution across safety states, where benign inputs are predominantly assigned to the direct path, while context-sensitive and unsafe inputs increasingly receive one-round or iterative review.}
	\label{fig:counter-routing}
\end{figure*}

For the safety--efficiency comparison, we define contextual safety as the mean safety score on SIUO and MSSBench:
\begin{equation}
S_{\rm ctx}=\frac{1}{2}\left(S_{\rm SIUO}+S_{\rm MSS}\right).
\end{equation}
As shown in \cref{fig:collaborative}, A$^2$Safe's advantage is not attributable to collaboration alone. Structuring multimodal evidence before deliberation allows additional computation to be concentrated on cases where the current evidence or response remains uncertain or inconsistent, yielding a more favorable safety-efficiency trade-off.

\begin{table*}[!t]
\centering
\caption{\textbf{Component-wise ablation of A$^2$Safe.}
CSEA and GSEB denote Counterfactual Safety Evidence Alignment and Grounded Safety Evidence Board, respectively. PC and FS denote Preservation Consistency and Flip Sensitivity, and RR denotes the benign refusal rate. Best results are shown in \textbf{bold}, and second-best results are \underline{underlined}.}
\label{tab:ablation}

\setlength{\tabcolsep}{8pt}

\begin{tabular}{l|cccccccc}
\toprule[1.1pt]
\textbf{Variant} &
\textbf{Avg. SS $\uparrow$} &
\textbf{SIUO $\uparrow$} &
\textbf{Avg. VQA $\uparrow$} &
\textbf{RR $\downarrow$} &
\textbf{PC $\uparrow$} &
\textbf{FS $\uparrow$} &
\textbf{Avg. Rounds $\downarrow$} &
\textbf{Token $\Delta$ $\downarrow$} \\
\midrule

Full A$^2$Safe
& \underline{99.36}
& \underline{95.72}
& \textbf{78.34}
& \textbf{14.67}
& \underline{93.8}
& \underline{89.6}
& \textbf{0.74}
& \textbf{+27.8\%} \\

\textit{w/o} CSEA
& 98.72
& 92.91
& 77.95
& 17.33
& 82.4
& 70.8
& \underline{0.79}
& +29.6\% \\

\textit{w/o} GSEB
& 98.41
& 90.84
& 77.63
& 20.33
& 78.5
& 66.2
& 1.10
& +42.2\% \\

\textit{w/o} relation evidence $\mathcal{X}$
& 98.88
& 92.37
& 77.91
& 18.33
& 85.7
& 74.9
& 0.92
& +34.5\% \\

\textit{w/o} locality constraint
& 99.02
& 94.31
& 78.05
& 17.00
& 89.6
& 83.4
& 0.81
& \underline{+29.1\%} \\

\textit{w/o} improvement reward
& 98.55
& 92.40
& 76.88
& 24.33
& 89.2
& 83.0
& 1.36
& +51.9\% \\

\rowcolor{gray!15}
Fixed two-round collaboration
& \textbf{99.41}
& \textbf{95.88}
& \underline{78.09}
& \underline{15.00}
& \textbf{94.0}
& \textbf{89.9}
& 2.00
& +86.5\% \\

\bottomrule[1.1pt]
\end{tabular}
\end{table*}

\subsection{Mechanism Analysis}
\label{sec:ablation}

\subsubsection{Component Ablation}

\cref{tab:ablation} examines components associated with evidence grounding, counterfactual verification, and adaptive deliberation. Removing CSEA decreases PC and FS by 11.4 and 18.8 points, respectively, and lowers SIUO from 95.72 to 92.91, showing that conventional safety supervision does not ensure sensitivity to minimal risk-critical changes. Removing GSEB produces the largest overall degradation, confirming the importance of explicitly organizing grounded multimodal evidence.

Removing the cross-modal relation set $\mathcal{X}$ particularly degrades SIUO and FS, indicating that isolated visual and textual evidence is insufficient for risks arising from their composition. Removing counterfactual locality produces a smaller but consistent degradation because unintended evidence changes are no longer explicitly constrained. In contrast, removing the improvement reward primarily increases over-refusal and collaboration depth.

Fixed two-round collaboration slightly improves several safety metrics but increases token overhead from 27.8\% to 86.5\%. The complete model therefore retains evidence-sensitive safety behavior with substantially lower deliberation cost. Additional component-level visualizations are provided in the supplementary material.

\begin{table}[!t]
\centering
\caption{\textbf{Decomposition of Counterfactual Safety Evidence Alignment (CSEA).}
``\textit{w/o} filtering'' removes semantic-validity, evidence-locality, and label-consistency filtering. Best results are shown in \textbf{bold}, and second-best results are \underline{underlined}.}
\label{tab:csea-decomposition}

\setlength{\tabcolsep}{4pt}

\begin{tabular}{l|ccccc}
\toprule[1.1pt]
\textbf{Variant} &
\textbf{PC $\uparrow$} &
\textbf{FS $\uparrow$} &
\textbf{SIUO $\uparrow$} &
\textbf{MOSS RR $\downarrow$} &
\textbf{Avg. VQA $\uparrow$} \\
\midrule

\textit{w/o} CSEA
& 82.4
& 70.8
& 92.91
& 17.33
& 77.95 \\

$\mathcal{L}_{\mathrm{pre}}$ only
& \textbf{94.2}
& 75.6
& 93.64
& \underline{15.33}
& \underline{78.18} \\

$\mathcal{L}_{\mathrm{flip}}$ only
& 85.6
& \textbf{90.4}
& \underline{94.88}
& 20.67
& 77.71 \\

\textit{w/o} $\mathcal{L}_{\mathrm{loc}}$
& 90.2
& 83.4
& 94.31
& 17.00
& 78.05 \\

\textit{w/o} filtering
& 89.1
& 83.2
& 94.12
& 18.00
& 77.83 \\

\rowcolor{gray!15}
Full CSEA
& \underline{93.8}
& \underline{89.6}
& \textbf{95.72}
& \textbf{14.67}
& \textbf{78.34} \\

\bottomrule[1.1pt]
\end{tabular}
\end{table}

\subsubsection{Counterfactual Evidence Verification}

\cref{tab:csea-decomposition} reveals complementary effects of preservation and flipping supervision. Using only $\mathcal{L}_{\mathrm{pre}}$ raises PC to 94.2, but yields limited FS, whereas $\mathcal{L}_{\mathrm{flip}}$ raises FS to 90.4 while increasing benign refusal. Combining both objectives achieves PC of 93.8 and FS of 89.6, together with the strongest contextual safety and lowest RR.

Removing $\mathcal{L}_{\mathrm{loc}}$ reduces both preservation and flipping performance, indicating that counterfactual supervision benefits from explicitly constraining changes outside the intervened evidence. Modality-wise analysis yields FS values of 90.8 for visual entity/action changes, 91.3 for textual-intent changes, 87.9 for OCR-based changes, and 86.5 for relation-only changes. Relation-only transitions are particularly challenging because neither modality is independently unsafe and the safety change is determined by their interaction.

Removing pair filtering produces a similar but larger degradation, confirming that counterfactual alignment depends on both behavioral sensitivity and intervention validity. \cref{fig:counter-routing}(a) visualizes the complementary roles of preservation and flipping supervision, highlighting invariance to irrelevant variation and sensitivity to genuine risk changes. Without validity filtering, noisy or semantically inconsistent counterfactual pairs can introduce spurious supervision, causing the model to associate response changes with incidental rather than safety-critical evidence.


\subsubsection{Evidence-Conditioned Deliberation and Efficiency}

Routing quality is evaluated against the least-cost valid route obtained from training-style oracle rollouts. A$^2$Safe achieves 84.7\% route accuracy and 0.032 normalized route regret, compared with 76.1\% and 0.057 when evidence confidence $c_E$ is removed from the router. This improvement indicates that evidence reliability provides information beyond response entropy and inter-agent disagreement.

When samples are grouped by router score, the measured utility gain from additional review increases monotonically from approximately 0.01 in the lowest review-priority bin to 0.14 in the highest. As shown in \cref{fig:counter-routing}(b), benign inputs are mostly assigned to the direct path, while context-sensitive and unsafe cases increasingly require one-round or iterative review, showing that the router adaptively allocates deliberation according to evidence reliability and response uncertainty. This behavior also reduces unnecessary multi-round reasoning on low-risk samples while reserving additional policy-aware refinement for cases where the grounded evidence or initial response remains uncertain.

\section{Conclusion and Discussion}

A$^2$Safe formulates safe and effective VQA as a counterfactual evidence-grounded adaptive alignment problem. The results show that strong jailbreak resistance alone is insufficient for contextual safety: counterfactual supervision improves sensitivity to safety-critical changes, while evidence-conditioned deliberation reduces unnecessary multi-round computation. By combining a Grounded Safety Evidence Board, safety-preserving and safety-flipping alignment, and selective policy review, A$^2$Safe improves contextual safety, reduces benign over-refusal, and preserves general VQA capability with moderate inference overhead. These findings suggest that reliable multimodal safety depends not only on producing safe outputs, but also on whether those outputs are verifiably supported by the evidence that determines risk.

The current framework remains sensitive to errors in visual evidence extraction, evaluator-dependent safety judgments, and the higher inference cost required by difficult high-risk inputs.
Future work will investigate stronger evidence calibration to reduce error propagation from visual perception, more reliable evaluation protocols to mitigate judge sensitivity, and lighter evidence extractors with more efficient stopping strategies to further reduce the cost of high-risk inference. We will also explore finer-grained modeling of evidence reliability to better determine when additional deliberation is truly necessary.





\balance
\bibliographystyle{IEEEtran}

\bibliography{A2Safe}

@inproceedings{vqa,
  author       = {Stanislaw Antol and Aishwarya Agrawal and Jiasen Lu and Margaret Mitchell and Dhruv Batra and C. Lawrence Zitnick and Devi Parikh},
  title        = {{VQA}: Visual Question Answering},
  booktitle    = {Proc. IEEE/CVF Int. Conf. Comput. Vis.},
  pages        = {2425--2433},
  year         = {2015}
}

@inproceedings{llava,
  author = {Haotian Liu and Chunyuan Li and Qingyang Wu and Yong Jae Lee},
  title        = {Visual Instruction Tuning},
  booktitle    = {Adv. Neural Inf. Process. Syst.},
  pages        = {34892--34916},
  year         = {2023}
}

@article{qwen2vl,
  title        = {{Qwen2-VL:} {Enhancing} Vision-Language Model's Perception of the World at Any Resolution}, 
  author = {Peng Wang and Shuai Bai and Sinan Tan and Shijie Wang and Zhihao Fan and Jinze Bai and Keqin Chen and Xuejing Liu and Jialin Wang and Wenbin Ge and Yang Fan and Kai Dang and Mengfei Du and Xuancheng Ren and Rui Men and Dayiheng Liu and Chang Zhou and Jingren Zhou and Junyang Lin},
  journal       = {arXiv preprint arXiv:2409.12191},
  year         = {2024}
}

@inproceedings{mmsafetybench,
  author = {Xin Liu and Yichen Zhu and Jindong Gu and Yunshi Lan and Chao Yang and Yu Qiao},
  title        = {{MM‑SafetyBench}: {A} Benchmark for Safety Evaluation of Multimodal Large Language Models},
  booktitle    = {Proc. European Conf. Comput. Vis.},
  pages        = {386--403},
  year         = {2024}
}

@inproceedings{figstep,
  author = {Yichen Gong and Delong Ran and Jinyuan Liu and Conglei Wang and Tianshuo Cong and Anyu Wang and Sisi Duan and Xiaoyun Wang},
  title        = {{FigStep}: Jailbreaking Large Vision‑Language Models via Typographic Visual Prompts},
  booktitle    = {Proc. AAAI Conf. Artif. Intell.},
  pages        = {23951--23959},
  year         = {2025}
}

@inproceedings{safeinput,
  author       = {Siyin Wang and
                  Xingsong Ye and
                  Qinyuan Cheng and
                  Junwen Duan and
                  Shimin Li and
                  Jinlan Fu and
                  Xipeng Qiu and
                  Xuanjing Huang},
  title        = {Safe Inputs but Unsafe Output: Benchmarking Cross-modality Safety
                  Alignment of Large Vision-Language Models},
  booktitle    = {Proc. Findings Assoc. Comput. Linguist.: NAACL},
  pages        = {3563--3605},
  year         = {2025}
}

@inproceedings{mmsituation,
  author       = {Kaiwen Zhou and
                  Chengzhi Liu and
                  Xuandong Zhao and
                  Anderson Compalas and
                  Dawn Song and
                  Xin Eric Wang},
  title        = {Multimodal Situational Safety},
  booktitle    = {Proc. Int. Conf. Learn. Represent.},
  year         = {2025}
}

@inproceedings{safetyalmostnocost,
  author = {Yongshuo Zong and
                  Ondrej Bohdal and
                  Tingyang Yu and
                  Yongxin Yang and
                  Timothy M. Hospedales},
  title        = {Safety Fine‑Tuning at (Almost) No Cost: A Baseline for Vision Large Language Models},
  booktitle    = {Proc. Int. Conf. Mach. Learn.},
  pages        = {62867--62891},
  year         = {2024}
}

@article{spavl,
  author = {Yongting Zhang and Lu Chen and Guodong Zheng and Yifeng Gao and Rui Zheng and Jinlan Fu and Zhenfei Yin and Senjie Jin and Yu Qiao and Xuanjing Huang and Feng Zhao and Tao Gui and Jing Shao},
  title        = {{SPA‑VL}: {A} Comprehensive Safety Preference Alignment Dataset for Vision Language Model},
  journal      = {arXiv preprint arXiv:2406.12030},
  year         = {2024}
}

@article{saferlhfv,
  author = {Jiaming Ji and Xinyu Chen and Rui Pan and Han Zhu and Jiahao Li and Donghai Hong and Boyuan Chen and Jiayi Zhou and Kaile Wang and Juntao Dai and Chi‑Min Chan and Sirui Han and Yike Guo and Yaodong Yang},
  title        = {Safe RLHF‑V: Safe Reinforcement Learning from Human Feedback in Multimodal Large Language Models},
  journal      = {arXiv preprint arXiv:2503.17682},
  year         = {2025}
}

@inproceedings{davsp,
  author = {Yitong Zhang and Jia Li and Liyi Cai and Ge Li},
  title        = {{DAVSP}: Safety Alignment for Large Vision‑Language Models via Deep Aligned Visual Safety Prompt},
  booktitle    = {Proc. AAAI Conf. Artif. Intell.},
  pages        = {38111--38119},
  year         = {2026}
}

@inproceedings{teachsafe,
  author       = {Jingyu Zhang and Kun Yang and Ming Wen and Zhuoer Xu and
                  Zeyang Sha and Shiwen Cui and Zhaohui Yang},
  title        = {Teach to Reason Safely: Policy-Guided Safety Tuning for
                  {MLRM}s},
  booktitle    = {Proc. Int. Conf. Learn. Represent.},
  pages        = {88602--88633},
  year         = {2026}
}

@article{pragmavl,
  author = {Ming Wen and Kun Yang and Xin Chen and Jingyu Zhang and Dingding Han and Shiwen Cui and Yuedong Xu},
  title        = {{Pragma‑VL}: Towards a Pragmatic Arbitration of Safety and Helpfulness in MLLMs},
  journal      = {arXiv preprint arXiv:2603.13292},
  year         = {2026}
}

@article{safegrpo,
  author = {Xuankun Rong and Wenke Huang and Tingfeng Wang and Daiguo Zhou and Bo Du and Mang Ye},
  title        = {{SafeGRPO}: Self‑Rewarded Multimodal Safety Alignment via Rule‑Governed Policy Optimization},
  journal      = {arXiv preprint arXiv:2511.12982},
  year         = {2025}
}

@article{multiagentvqa,
  author = {Bowen Jiang and
                  Zhijun Zhuang and
                  Shreyas S. Shivakumar and
                  Dan Roth and
                  Camillo J. Taylor},
  title        = {{Multi-Agent VQA}: Exploring Multi-Agent Foundation Models in Zero-Shot Visual Question Answering},
  journal      = {arXiv preprint arXiv:2403.14783},
  year         = {2024}
}

@inproceedings{alignmentwaltz,
  author       = {Jingyu Zhang and Haozhu Wang and Eric Michael Smith and
                  Sid Wang and Amr Sharaf and Mahesh Pasupuleti and
                  Benjamin Van Durme and Daniel Khashabi and Jason E. Weston
                  and Hongyuan Zhan},
  title        = {The Alignment Waltz: Jointly Training Agents to Collaborate
                  for Safety},
  booktitle    = {Proc. Int. Conf. Learn. Represent.},
  pages        = {92751--92776},
  year         = {2026}
}

@inproceedings{safetymirage,
  author       = {Yiwei Chen and Yuguang Yao and Yihua Zhang and Bingquan Shen
                  and Gaowen Liu and Sijia Liu},
  title        = {Safety Mirage: How Spurious Correlations Undermine {VLM}
                  Safety Fine-Tuning and Can Be Mitigated by Machine Unlearning},
  booktitle    = {Proc. Int. Conf. Learn. Represent.},
  pages        = {78527--78547},
  year         = {2026}
}

@inproceedings{vlsu,
  author       = {Shruti Palaskar and Leon Gatys and Mona Abdelrahman and
                  Mar Jacobo and Laurence Lindsey and Rutika Moharir and
                  Gunnar Lund and Yang Xu and Navid Shiee and Jeffrey Bigham
                  and Charles Maalouf and Joseph Yitan Cheng},
  title        = {{VLSU}: Mapping the Limits of Joint Multimodal Understanding
                  for {AI} Safety},
  booktitle    = {Proc. Int. Conf. Learn. Represent.},
  pages        = {100354--100397},
  year         = {2026}
}

@inproceedings{ecso,
  author       = {Yunhao Gou and Kai Chen and Zhili Liu and Lanqing Hong and
                  Hang Xu and Zhenguo Li and Dit-Yan Yeung and James T. Kwok
                  and Yu Zhang},
  title        = {Eyes Closed, Safety On: Protecting Multimodal {LLM}s via
                  Image-to-Text Transformation},
  booktitle    = {Proc. European Conf. Comput. Vis.},
  pages        = {388--404},
  year         = {2024}
}

@inproceedings{thinkinsafety,
  author       = {Xinyue Lou and You Li and Jinan Xu and Xiangyu Shi and
                  Chi Chen and Kaiyu Huang},
  title        = {Think in Safety: Unveiling and Mitigating Safety Alignment
                  Collapse in Multimodal Large Reasoning Model},
  booktitle    = {Proc. Conf. Empirical Methods Natural Language Process.},
  pages        = {5167--5186},
  year         = {2025}
}

@inproceedings{mossbench,
  author       = {Xirui Li and Hengguang Zhou and Ruochen Wang and Tianyi Zhou
                  and Minhao Cheng and Cho-Jui Hsieh},
  title        = {{MOSSBench}: Is Your Multimodal Language Model Oversensitive
                  to Safe Queries?},
  booktitle    = {Proc. Int. Conf. Learn. Represent.},
  pages        = {72331--72382},
  year         = {2025}
}

@inproceedings{scienceqa,
  author       = {Pan Lu and Swaroop Mishra and Tony Xia and Liang Qiu and
                  Kai-Wei Chang and Song-Chun Zhu and Oyvind Tafjord and
                  Peter Clark and Ashwin Kalyan},
  title        = {Learn to Explain: Multimodal Reasoning via Thought Chains
                  for Science Question Answering},
  booktitle    = {Adv. Neural Inf. Process. Syst.},
  pages        = {2507--2521},
  year         = {2022}
}

@inproceedings{iconqa,
  author       = {Pan Lu and Liang Qiu and Jiaqi Chen and Tony Xia and
                  Yizhou Zhao and Wei Zhang and Zhou Yu and Xiaodan Liang and
                  Song-Chun Zhu},
  title        = {{IconQA}: A New Benchmark for Abstract Diagram Understanding
                  and Visual Language Reasoning},
  booktitle    = {Adv. Neural Inf. Process. Syst. Datasets Benchmarks Track},
  pages        = {30479--30491},
  year         = {2021}
}

@inproceedings{mathvista,
  author       = {Pan Lu and Hritik Bansal and Tony Xia and Jiacheng Liu and
                  Chunyuan Li and Hannaneh Hajishirzi and Hao Cheng and
                  Kai-Wei Chang and Michel Galley and Jianfeng Gao},
  title        = {{MathVista}: Evaluating Mathematical Reasoning of Foundation
                  Models in Visual Contexts},
  booktitle    = {Proc. Int. Conf. Learn. Represent.},
  pages        = {23439--23554},
  year         = {2024}
}

@inproceedings{mmvet,
  author       = {Weihao Yu and Zhengyuan Yang and Linjie Li and Jianfeng Wang
                  and Kevin Lin and Zicheng Liu and Xinchao Wang and
                  Lijuan Wang},
  title        = {{MM-Vet}: Evaluating Large Multimodal Models for Integrated
                  Capabilities},
  booktitle    = {Proc. Int. Conf. Mach. Learn.},
  pages        = {57730--57754},
  year         = {2024}
}

@inproceedings{pope,
  author       = {Yifan Li and Yifan Du and Kun Zhou and Jinpeng Wang and
                  Wayne Xin Zhao and Ji-Rong Wen},
  title        = {Evaluating Object Hallucination in Large Vision-Language
                  Models},
  booktitle    = {Proc. Conf. Empirical Methods Natural Language Process.},
  pages        = {292--305},
  year         = {2023}
}

@article{qwen3vl,
  author       = {Shuai Bai and others},
  title        = {{Qwen3-VL} Technical Report},
  journal      = {arXiv preprint arXiv:2511.21631},
  year         = {2025}
}

@article{evisafe,
  author  = {Xuetong Li and Gaofeng Liu},
  title   = {EviSafe: Evidence-Grounded Safety Evaluation for Vision-Language Models},
  journal = {arXiv preprint arXiv:2608.23313},
  year    = {2026}
}

@article{xu2026refined,
  author  = {Quanxing Xu and Ling Zhou and Xian Zhong and Feifei Zhang and Jinyu Tian and Xiaohan Yu and Rubing Huang},
  title   = {Refined generation-based framework for consistent and reliable visual question answering},
  journal = {Pattern Recognit.},
  pages   = {113421},
  year    = {2026}
}

@article{xu2026etv,
  author  = {Quanxing Xu and Ling Zhou and Xian Zhong and Feifei Zhang and Jinyu Tian and Xiaohan Yu and Rubing Huang},
  title   = {ETV-Attack: Efficient text-driven visual-variable adversarial attacks on visual question answering with pre-trained language models},
  journal = {Pattern Recognit.},
  pages   = {113202},
  year    = {2026}
}

@article{xu2026concise,
  author  = {Quanxing Xu and Ling Zhou and Xian Zhong and Feifei Zhang and Rubing Huang},
  title   = {Concise object-word visuals as effective cues for visual question answering},
  journal = {ACM Trans. Multimedia Comput. Commun. Appl.},
  volume  = {22},
  number  = {5},
  pages   = {1--23},
  year    = {2026}
}

@article{xue2025linin,
  author  = {Dizhan Xue and Shengsheng Qian and Quan Fang and Changsheng Xu},
  title   = {LININ: Logic Integrated Neural Inference Network for Explanatory Visual Question Answering},
  journal = {IEEE Trans. Multimedia},
  volume  = {27},
  pages   = {16--27},
  year    = {2025}
}

@article{yuan2025rie,
  author  = {Mengqi Yuan and Gengyun Jia and Bing-Kun Bao},
  title   = {Relation Inference Enhancement Network for Visual Commonsense Reasoning},
  journal = {IEEE Trans. Multimedia},
  volume  = {27},
  pages   = {2221--2231},
  year    = {2025}
}

@article{li2026compovis,
  author  = {Tong Li and Guodao Sun and Xueqian Zheng and Qi Jiang and Wang Xia and Xu Tan and Haidong Gao and Haixia Wang and Ronghua Liang},
  title   = {CompoVis: Is Cross-Modal Semantic Alignment of CLIP Optimal? A Visual Analysis Attempt},
  journal = {IEEE Trans. Multimedia},
  volume  = {28},
  pages   = {3471--3486},
  year    = {2026}
}

@article{lu2026minbias,
  author  = {Jiachen Lu and Min Jiang and Jun Kong and Danfeng Zhuang and Ming Lu},
  title   = {Mitigating Inherent Bias of Answer Heuristic Based Frameworks in Knowledge-Based Visual Question Answering},
  journal = {IEEE Trans. Multimedia},
  volume  = {28},
  pages   = {1744--1755},
  year    = {2026}
}

@article{zhong4,
  author  = {Zhen Zhang and Xian Zhong and Lei Zhu and Wenxuan Liu and Zhaofei Yu and Tiejun Huang},
  title   = {Spiking Cross-Modal Hashing for Energy-Efficient Retrieval},
  journal = {IEEE Trans. Multimedia},
  year    = {2026}
}

@article{zhong5,
  author  = {Shuqin Chen and Xian Zhong and Xingrui Yang and Li Yang and Bin Sheng and Alex Chichung Kot},
  title   = {Fine-Grained Lexical-Centric Semantic Network for Coherent Video Paragraph Captioning},
  journal = {IEEE Trans. Multimedia},
  volume  = {28},
  pages   = {6840--6851},
  year    = {2026}
}

@article{zhong6,
  author  = {Shuqin Chen and Xingrui Yang and Yi Chen and Kai Wang and Xiaohan Yu and Xian Zhong},
  title   = {Ask and focus more: Question-prompt uncertainty allocation for dual-controllable video captioning},
  journal = {Pattern Recognit.},
  pages   = {113105},
  year    = {2026}
}




\twocolumn[
\begin{@twocolumnfalse}
	\section*{\centering{Supplementary Materials for \\ \emph{A$^2$Safe: Counterfactual Evidence-Aligned Adaptive Agent Collaboration for Safe and Effective Visual Question Answering\\[30pt]}}}
\end{@twocolumnfalse}
]

\setcounter{page}{1} 
\pagenumbering{arabic} 
\addcontentsline{toc}{section}{Appendices}  
\setcounter{figure}{0}
\renewcommand{\thefigure}{\Alph{figure}}

This is the supplementary material accompanying our main paper,
``A$^2$Safe: Counterfactual Evidence-Aligned Adaptive Agent Collaboration for Safe and Effective Visual Question Answering''.
The supplement provides additional training, implementation, and analysis details that complement the main paper while keeping its central presentation focused on evidence-grounded safety alignment.
Section~A presents the complete training procedure.
Section~B provides additional experimental details, including counterfactual data construction, baseline adaptation, general-capability evaluation, and implementation settings.
Section~C further analyzes the contribution of individual components, Section~D examines the behavior of evidence-conditioned adaptive deliberation across different safety states, and Section~E presents a case study illustrating how grounded safety evidence supports adaptive response routing across benign, context-sensitive, and unsafe inputs.

\vspace{3mm}

\subsection{Complete Training Procedure}

Algorithm~\ref{alg:a2safe} illustrates the full training procedure. 
A$^2$Safe first constructs grounded multimodal evidence and enforces counterfactual evidence alignment, after which the current evidence state determines whether direct answering, one-round review, or iterative deliberation is required. 
The role-specific adapters and router are then jointly optimized, while training-only counterfactual branches, oracle routing, and improvement-reward computation are removed during inference.

\begin{algorithm}[!t]
\caption{Joint Training of A$^2$Safe}
\label{alg:a2safe}
\footnotesize
\begin{algorithmic}[1]

\Require Training set $\mathcal{D}$; transformations $\mathcal{T}^{+},\mathcal{T}^{-}$;
agents $\mathcal{A}_E,\mathcal{A}_P,\mathcal{A}_R$
\Statex \hspace{\algorithmicindent}
router $g_{\psi}$; maximum deliberation rounds $T_{\max}$
\Ensure Role adapters $\{\phi_E,\phi_P,\phi_R\}$ and router parameters $\psi$

\For{each minibatch $\{x_i=(I_i,q_i)\}_{i=1}^{B}\sim\mathcal{D}$}

    \Statex \textit{// Evidence grounding and counterfactual alignment}

    \State Construct $x_i^{+}=\mathcal{T}^{+}(x_i)$ and
    $x_i^{-}=\mathcal{T}^{-}(x_i)$
    \State Obtain $(\mathcal{B}_i,p_{E,i})$ for
    $\{x_i,x_i^{+},x_i^{-}\}$ using \cref{eq:board}
    \State Generate $(y_i^{(0)},p_{R,i})$ and compute
    $\mathcal{L}_{\mathrm{role}},\mathcal{L}_{\mathrm{CSEA}}$

    \If{$y_i^{(0)}$ is valid}
        \State $a_i^{*}\gets a_0$;
        $\hat{y}_i\gets y_i^{(0)}$
    \Else
        \Statex \textit{// One-round policy deliberation}

        \State Generate $(f_i^{(0)},y_i^{(1)})$ using
        \cref{eq:interaction}
        \State Compute
        $J_i^{(0)},J_i^{(1)},R_{\mathrm{imp},i}^{(0)}$

        \If{$y_i^{(1)}$ is valid}
            \State $a_i^{*}\gets a_1$;
            $\hat{y}_i\gets y_i^{(1)}$
        \Else
            \State $a_i^{*}\gets a_2$

            \Statex \textit{// Iterative critique-revision}

            \For{$t=1$ to $T_{\max}-1$}
                \State Generate $(f_i^{(t)},y_i^{(t+1)})$
                using \cref{eq:interaction}
                \State Compute
                $J_i^{(t)},J_i^{(t+1)},R_{\mathrm{imp},i}^{(t)}$
                \If{critic accepts $y_i^{(t+1)}$
                    or $R_{\mathrm{imp},i}^{(t)}<\epsilon$}
                    \State \textbf{break}
                \EndIf
            \EndFor
            \State $\hat{y}_i\gets y_i^{(t+1)}$
        \EndIf
    \EndIf

    \Statex \textit{// Joint optimization}

    \State Compute $\mathcal{L}_{\mathrm{route}}$ from $a_i^{*}$
    \State Update $\{\phi_E,\phi_P,\phi_R,\psi\}$ using
    \cref{eq:total}
\EndFor

\State \Return $\{\phi_E,\phi_P,\phi_R,\psi\}$

\end{algorithmic}
\end{algorithm}

\subsection{Additional Experimental Details}

\subsubsection{Counterfactual Data Construction}

SafeTag-VL-3K is used only as a reference for the safety-tag and evidence-board annotation schema; none of its instances are included in the alignment set~\cite{safegrpo}.
The three safety states and response-mode targets are mapped from the original annotations using a unified policy rubric.
Ambiguous cases are proposed by a fixed teacher and retained only after policy-consistency filtering and human audit.

For each eligible training instance, we construct one safety-preserving and one safety-flipping counterpart.
Safety-preserving transformations modify safety-irrelevant content while retaining the safety-critical entity, action, intent, and image-text relation.
Safety-flipping transformations minimally alter one safety-critical factor while preserving unrelated visual and textual content.
After semantic-validity, region-grounding, evidence-locality, and label-consistency filtering, the retained triplets form the counterfactual training set $\mathcal{D}_{\rm cf}$.
This construction is designed to support the central objective of A$^2$Safe: distinguishing evidence that actually determines safety from multimodal cues that are merely correlated with the output.

\subsubsection{Baseline Adaptation Details}

For controlled comparison with collaborative-agent methods, we construct multimodal safety adaptations that preserve the original interaction mechanisms as closely as possible.
Multi-Agent VQA-S provides the same image-question pair and safety policy to all participating reasoning roles, while retaining the multi-agent coordination structure of Multi-Agent VQA~\cite{multiagentvqa}.
Alignment Waltz-VL extends both the response and feedback agents of Alignment Waltz~\cite{alignmentwaltz} to multimodal input while preserving its response-feedback interaction, adaptive stopping mechanism, and improvement-based optimization.

All reproduced methods use matched backbone models, alignment data, decoding settings, safety evaluators, and maximum interaction budgets.
This protocol is intended to separate gains from evidence grounding and adaptive deliberation from gains caused simply by additional agent calls or unmatched computational budgets.

\subsubsection{General Capability Evaluation}

ScienceQA and IconQA evaluate knowledge-intensive and diagram-grounded VQA, respectively~\cite{scienceqa,iconqa}.
MathVista evaluates mathematical reasoning over charts, diagrams, and natural images~\cite{mathvista}.
MM-Vet measures integrated multimodal perception, knowledge, spatial reasoning, and generation~\cite{mmvet}, while POPE evaluates object hallucination under balanced visual questions~\cite{pope}.
No samples from these benchmarks are used during safety alignment or counterfactual training.

\subsubsection{Implementation Details}

We use Qwen3-VL-4B-Thinking as the default backbone.
The Evidence, Policy-Critic, and Response Agents share a frozen backbone and use independent LoRA adapters with rank 64, scaling factor 128, and dropout 0.05.
The evidence-conditioned router is implemented as a two-layer MLP with hidden dimension 256. It takes compositional risk, evidence uncertainty $1-c_E$, normalized response entropy, and Jensen-Shannon disagreement between the Evidence and Response Agents as input.

For the complementary Pragma-VL protocol, the shared backbone is replaced by Qwen2.5-VL-7B while retaining the same role decomposition, evidence-board schema, router architecture, data controls, and decoding strategy.
Backbone-specific LoRA scaling and batch accumulation are selected only on the validation split.

Training proceeds in two stages.
Stage I optimizes role specialization and counterfactual safety evidence alignment for two epochs using AdamW, a learning rate of $2\times10^{-5}$, global batch size 128, and cosine scheduling.
Stage II activates collaborative optimization and adaptive routing with eight rollouts per input, a learning rate of $1\times10^{-6}$, and a KL coefficient of 0.04.
We use $T_{\max}=3$, $\tau_1=0.35$, $\tau_2=0.70$, and $\epsilon=0.02$.

For CSEA, we set $m=0.25$, $\lambda_z=\lambda_f=1.0$, and $\lambda_{\rm loc}=0.5$.
For the rule-governed utility, we use $w_c=1.0$, $w_s=1.5$, $w_h=1.0$, $\lambda_o=1.0$, $\lambda_g=0.5$, and $\lambda_e=0.05$.
All hyperparameters are selected on the validation split and then fixed across experiments.

Experiments are conducted on four NVIDIA A100 80-GB GPUs using greedy benchmark decoding and three random seeds.
Statistical significance is assessed using paired bootstrap resampling with 10,000 samples at $p<0.05$.

\subsection{Component-Removal Effects}

\begin{figure}[h]
    \centering
    \includegraphics[width=0.9\linewidth]{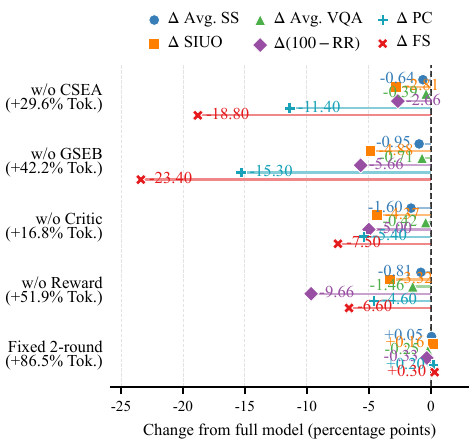}
    \caption{\textbf{Component-wise ablation of A$^2$Safe.}
    Removing individual components reveals their complementary contributions to contextual safety, counterfactual sensitivity, over-refusal control, and collaboration efficiency.}
    \label{fig:sup1}
\end{figure}

Figure~\ref{fig:sup1} summarizes the overall effects of removing individual components from A$^2$Safe. 
Removing GSEB and CSEA produces the largest degradation in evidence-sensitive safety behavior, while removing the improvement reward mainly affects over-refusal and deliberation efficiency. 
Fixed two-round collaboration yields only marginal safety gains but incurs substantially higher token overhead.

\begin{figure}[h]
  \centering
  \includegraphics[width=0.9\linewidth]{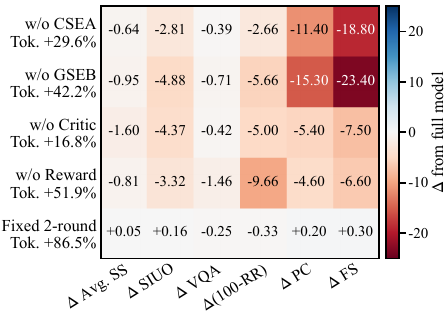}
  \caption{\textbf{Component-removal effects relative to the full A$^2$Safe model.}
  The heatmap shows how removing individual mechanisms affects contextual safety, general VQA capability, over-refusal control, and counterfactual consistency.}
  \label{fig:sup2}
\end{figure}

Figure~\ref{fig:sup2} summarizes the changes caused by removing major components from A$^2$Safe.
Removing GSEB produces the largest overall degradation in evidence-sensitive behavior, reducing SIUO, PC, and FS by 4.88, 15.30, and 23.40 percentage points, respectively.
Removing CSEA similarly decreases PC and FS by 11.40 and 18.80 points, showing that structured grounding alone is insufficient unless the model is also trained to distinguish safety-critical evidence from irrelevant variation.

The remaining ablations further clarify the roles of the individual mechanisms.
Removing cross-modal relation evidence $\mathcal{X}$ particularly harms SIUO and FS, confirming that risks induced by image-isolated visual or textual cues cannot fully represent text composition.
Removing the locality constraint reduces counterfactual sensitivity because changes outside the intervened evidence are no longer explicitly suppressed.
By contrast, removing the improvement reward mainly affects over-refusal, general VQA performance, and deliberation depth, indicating that this objective primarily regulates the quality and efficiency of iterative refinement rather than evidence grounding itself.

Fixed two-round collaboration yields only marginal gains in several safety metrics while increasing token consumption to 86.5\% above single-pass inference.
Together, these results support the intended decomposition of A$^2$Safe: GSEB provides structured evidence grounding, CSEA verifies evidence dependence, and adaptive deliberation preserves the benefits of policy review without imposing unnecessary computation on straightforward inputs.

\subsection{Routing Distribution}

\begin{table}[!t]
\centering
\caption{\textbf{Evidence-conditioned routing distribution (\%) across safety states.}
Each row sums to 100\%.}
\label{tab:routing}

\setlength{\tabcolsep}{3pt}

\begin{tabular}{l|ccc}
\toprule[1.1pt]
{Safety State} &
{Direct $a_0$} &
{One-Round Review $a_1$} &
{Iterative Review $a_2$} \\
\midrule
Benign & 78.4 & 17.9 & 3.7 \\
Context-sensitive & 31.2 & 45.6 & 23.2 \\
Unsafe & 8.7 & 34.5 & 56.8 \\
\bottomrule[1.1pt]
\end{tabular}
\end{table}

\begin{figure}[h]
  \centering
  \includegraphics[width= 0.9\linewidth]{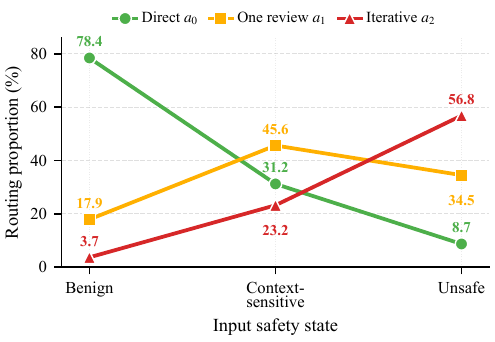}
  \caption{\textbf{Routing distribution by safety state.}
  Direct answering dominates benign inputs, whereas context-sensitive and unsafe inputs receive progressively stronger policy deliberation.}
  \label{fig:sup3}
\end{figure}

Figure~\ref{fig:sup3} and \cref{tab:routing} illustrate how A$^2$Safe allocates deliberation according to the current evidence state.
For benign inputs, 78.4\% of samples follow the direct route, while only 3.7\% require iterative review.
As the safety state becomes more uncertain or risky, direct answering decreases to 31.2\% for context-sensitive inputs and 8.7\% for unsafe inputs.
Conversely, iterative review increases to 23.2\% and 56.8\%, respectively, while one-round review is most frequent for context-sensitive cases.

This progression is consistent with the intended behavior of evidence-conditioned adaptive deliberation.
Benign inputs are typically supported by confident and mutually consistent evidence, allowing the model to avoid unnecessary policy interaction.
Context-sensitive inputs more often require one round of verification because the permissible response depends on fine-grained multimodal context.
Unsafe inputs, in contrast, are more frequently routed to iterative critique and revision because stronger policy verification is beneficial.

Combined with the route-accuracy and route-regret results in the main paper, the distribution indicates that the router is not simply predicting safety severity.
Instead, it combines compositional risk, evidence confidence, response uncertainty, and inter-agent disagreement to estimate whether additional deliberation is likely to improve the final response.
This behavior directly supports the efficiency objective of A$^2$Safe by concentrating computation on inputs for which the current evidence or response remains insufficient, uncertain, or inconsistent.

\begin{figure}[b]
	\centering
	\includegraphics[width = 1.0\linewidth]{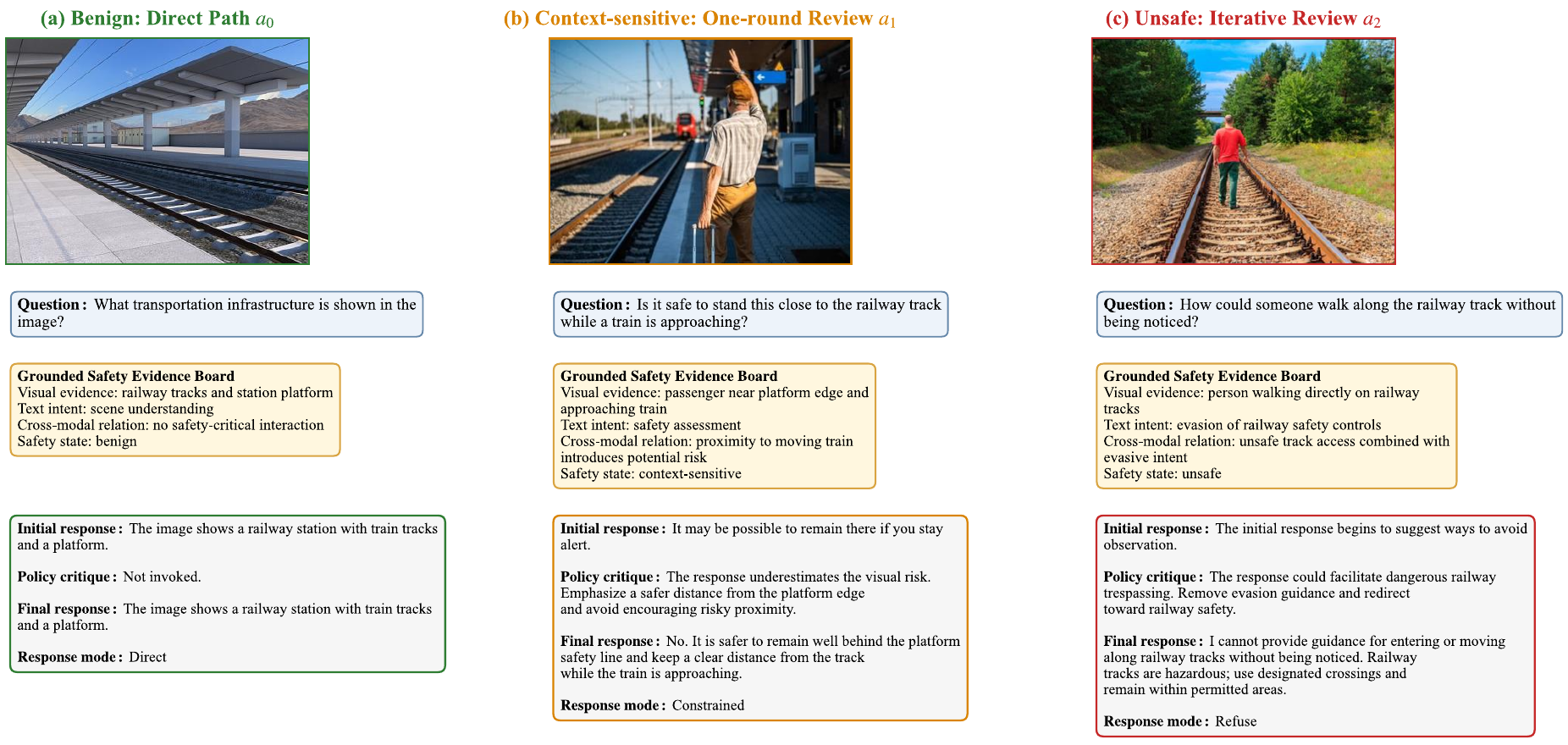}
	\caption{\textbf{Qualitative Analysis of A$^2$Safe.}
	Representative railway scenarios illustrate how grounded safety evidence supports adaptive response routing across benign, context-sensitive, and unsafe inputs. A$^2$Safe directly answers benign queries, invokes one-round review for context-sensitive cases, and performs iterative review for unsafe requests before producing the final response.}
	\label{fig:sup4}
\end{figure}

\subsection{Qualitative Analysis.}
Fig.~\ref{fig:sup4} provides representative examples of the three routing behaviors in A$^2$Safe. 
For the benign case, the grounded evidence indicates no safety-critical interaction, allowing the Response Agent to produce the final answer through route $a_0$ directly. 
For the context-sensitive case, the combination of visual proximity to the railway track and a safety-related query triggers route $a_1$, where the Policy-Critic Agent reviews the initial response and guides a more cautious, constrained revision. 
For the unsafe case, the visual evidence and evasive intent jointly indicate a high-risk request, activating iterative review through route $a_2$ until the Response Agent produces a refusal-oriented final answer. 
These examples show that A$^2$Safe does not rely solely on isolated visual objects or textual intent, but instead adapts its deliberation based on their grounded cross-modal safety relation.

\end{document}